\documentclass{egpubl}
\preprint
\usepackage[T1]{fontenc}
\usepackage{dfadobe}
\usepackage{cite}
\BibtexOrBiblatex
\electronicVersion
\PrintedOrElectronic
\usepackage{graphicx}
\ifpdf \pdfcompresslevel=9 \fi
\usepackage{egweblnk}
\usepackage{amsmath,amssymb}
\usepackage{pgfplots}
\pgfplotsset{compat=1.18}
\usepackage{stfloats}
\usepackage{placeins}
\usepackage{etoolbox}
\makeatletter
\copyrightTextTitPag{}
\copyrightTextRunPag{}
\def\ps@titlepage{%
  \let\@mkboth\@gobbletwo
  \def\@oddhead{}\let\@evenhead\@oddhead
  \def\@oddfoot{\hfil\thepage\hfil}\let\@evenfoot\@oddfoot
  \let\sectionmark\EmptySectionmark
  \let\subsectionmark\EmptySubsectionmark
}
\patchcmd{\@maketitle}{\vskip 35pt}{\vskip 16pt}{}{}
\patchcmd{\@maketitle}{\vskip 46pt}{\vskip 18pt}{}{}
\makeatother
\hypersetup{%
  pdfsubject={Learned real-time supersampling and temporal consistency},
  pdfkeywords={temporal supersampling, temporal consistency, sample registration, real-time rendering},
  colorlinks=true,linkcolor=blue,citecolor=blue,urlcolor=blue
}
\AtBeginDocument{\hypersetup{%
  pdfauthor={Hao Yu, Guoying Zhao},
  pdftitle={Frozen History: When Temporal Losses Stop Accumulation in Learned Real-Time Supersampling}
}}
\apptocmd{\maketitle}{\hypersetup{%
  pdfauthor={Hao Yu, Guoying Zhao},
  pdftitle={Frozen History: When Temporal Losses Stop Accumulation in Learned Real-Time Supersampling}
}}{}{}

\usepackage{booktabs}
\usepackage{xcolor}
\usepackage{tikz}
\usetikzlibrary{arrows.meta,positioning,fit,calc,backgrounds,shapes.misc}
\graphicspath{{figures/}}

\title[Frozen History]%
      {Frozen History: When Temporal Losses Stop Accumulation\\in Learned Real-Time Supersampling}

\author[H. Yu and G. Zhao]{%
Hao Yu$^{1,2}$ and Guoying Zhao$^{1,2,*}$\\
$^{1}$ELLIS Institute Finland\\
$^{2}$CMVS, University of Oulu, Finland\\
\href{mailto:hao.2.yu@oulu.fi}{hao.2.yu@oulu.fi},
\href{mailto:guoying.zhao@oulu.fi}{guoying.zhao@oulu.fi}\\
$^{*}$Corresponding author: Guoying Zhao
}

\begin{document}

%% ======== FIGURE 1 (Q1_teaser) ========
\teaser{%
  \centering
  \includegraphics[width=\linewidth]{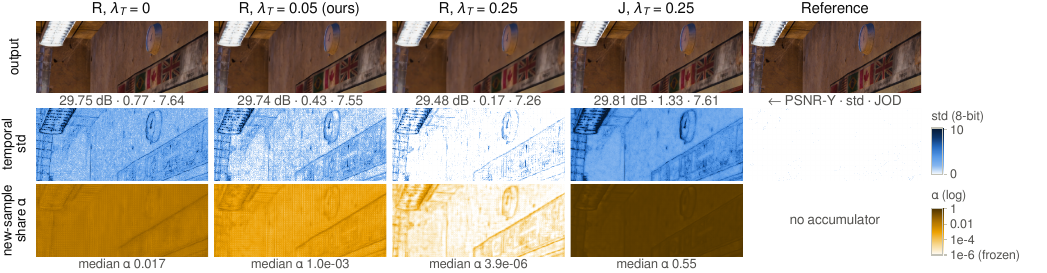}
  \caption{Correct registration makes frozen history inexpensive under the temporal loss. A static view (AbandonedSchool, $\times$4) reconstructed by our 192-channel network (45k iterations) trained with registered (R) or jitter-carrying (J) motion and the weight above each column. Rows: output, per-pixel temporal standard deviation and deployed share $\alpha$ of the newest sample (log scale; light: history copied) over the static segment; the strip gives PSNR-Y (test set), standard deviation (this segment) and JOD (three static segments), and the label under each $\alpha$ map its median in this crop. Under R with $\lambda_T=0.25$ the network stops accumulating: its flicker is lowest, but it blurs the wall texture and flags (1.1\,dB below $\lambda_T=0.05$ in this crop) and scores lowest in ColorVideoVDP. Under J the same loss neither freezes nor stabilises.}
  \label{fig:teaser}%
}

\maketitle

\begin{abstract}
%% ======== BEGIN TEXT 00_abstract ========
Learned real-time supersampling reconstructs high-resolution frames by accumulating jittered low-resolution samples along motion vectors, and many methods add a temporal consistency loss against flicker with a hand-tuned weight. We show that with accurate registration, this loss can suppress accumulation: on static content the network copies its warped history at almost no temporal cost. Motion vectors that carry the camera jitter make this frozen history expensive and can hide the effect. Exact registration is itself easy to get wrong with renderer buffers; we identify four errors whose effects training can conceal: motion vectors that include the jitter, one motion field for both the low- and high-resolution lattices, jittered depth and an unregistered skip path. A model-free oracle shows that registration makes freezing two orders of magnitude cheaper on static pixels, and controlled training runs show that correct registration reverses the effect of the temporal weight on PSNR: under exact registration, increasing the temporal weight reduces pixel variance at the expense of reconstruction accuracy and perceived quality, with a 0.7--0.9\,dB loss on static frames at a weight of 0.25. The reversal recurs in three network families and at two scales; on the unjittered STSS dataset the loss also costs accuracy. With a small temporal weight and a normalised sample accumulator, a plain registered network improves the best published PSNR on the QRISP benchmark by 0.34 to 0.62\,dB at $\times$2, $\times$3 and $\times$4 upscaling to 1080p. At $\times$2, its pixel standard deviation on the QRISP static segment is one quarter that of DLSS~2. Without retraining, it leads classical temporal upsampling by at least 4.4\,dB on Arm's jittered Bistro capture. Our most accurate $\times$4 model runs in 2.03\,ms per 1080p frame on an H200.
\newline Code: \url{https://github.com/CV-AC/Frozen_history}.
%% ======== END TEXT 00_abstract ========
%-------------------------------------------------------------------------
\end{abstract}

%-------------------------------------------------------------------------
\twocolumn  % Start the introduction on a fresh page with both columns reset.
\section{Introduction}\label{sec:intro}
%% ======== BEGIN TEXT 01_introduction ========
Interactive rendering in games must produce a detailed image within a frame budget shared by shading, geometry and post-processing. Temporal supersampling saves shading work by rendering below the display resolution and reconstructing a full-resolution image from samples gathered over successive frames~\cite{yang2020survey,mercier2023efficient,liu2020dlss}. At $\times$4 upscaling in each dimension, the renderer shades one input pixel for every 16 output pixels; the reconstruction must recover detail from this sparse input while leaving time for rendering. Temporal anti-aliasing and upsampling use sub-pixel camera jitter to obtain complementary samples, then align and accumulate them using renderer motion vectors~\cite{yang2009amortized,karis2014high,yang2020survey}. Learned methods combine current samples with reprojected history using a network trained against a high-resolution reference~\cite{mercier2023efficient,liu2020dlss}.

The value of accumulation is especially clear when the camera stops. The view stays fixed, but jitter continues to sample different positions. Incorporating these samples should recover edges and fine texture that a single low-resolution frame cannot resolve. An estimate that stops changing can also stop improving. We study how temporal training losses suppress sample updates.

Training with per-frame reconstruction alone can leave visible flicker, so many methods add a temporal consistency term that matches output changes to reference changes~\cite{chaitanya2017interactive,thomas2022temporally,herveau2023minimal}. Its weight is usually chosen empirically: Thomas et al.~\shortcite{thomas2022temporally} adjust the spatial contribution to balance stability and sharpness, while Herveau et al.~\shortcite{herveau2023minimal} give the temporal term the greatest influence. We examine this choice together with sample registration. With accurate registration, a static reference changes little after reprojection. Copying the previous output therefore incurs almost no temporal penalty and approximately preserves its existing reconstruction error. Continued accumulation can introduce fluctuations as new jittered samples enter the estimate, which the temporal term penalises. A sufficiently large weight can favour retaining an imperfect history over incorporating new information. We call this behaviour \emph{frozen history}; its temporal penalty is the reference's reprojected change, or \emph{copy cost} (Section~\ref{sec:mechanism}).

The copy cost depends on registration. Motion vectors that include camera jitter displace the history even on static content, raising the copy cost and masking the failure. On QRISP~\cite{mercier2023efficient}, a model-free oracle shows that using the dataset's jitter-carrying motion vectors without compensation makes freezing 118 times more expensive on static pixels. Correct registration also requires distinguishing the fixed output lattice from the moving lattice of the rendered samples. We identify four errors in how renderer buffers are used and register each state on its own lattice (Section~\ref{sec:registration}). Training can conceal these errors through learned compensation.

Controlled experiments confirm this behaviour (Figure~\ref{fig:teaser}). With exact registration, larger temporal weights monotonically reduce static pixel variance while lowering accuracy. At $\lambda_T=0.25$, our $\times$4 network's median new-sample share $\alpha$ falls below $10^{-4}$ on the School stop, and PSNR over the 97 static-camera frames drops by 0.86\,dB relative to $\lambda_T=0$. With jitter-carrying motion, the same loss has little effect on static flicker. Registration reverses the direction of the PSNR response across three network families, including one without an explicit accumulator, and at two scales (Sections~\ref{sec:dial} and~\ref{sec:generality}). A larger temporal weight also lowers accuracy on STSS~\cite{he2024low}, whose inputs are rendered without jitter.

We therefore use a small temporal weight, accepting a modest loss of static accuracy for less flicker, and a normalised sample accumulator that improves stability at comparable accuracy. We also report reconstruction and perceptual quality: both pixel variance and temporal PSNR reward freezing. Our contributions are:
\begin{itemize}
\item We derive the copy cost of a frozen output and verify, with the oracle and controlled training, how registration allows the temporal loss to suppress accumulation on static content (Section~\ref{sec:mechanism}).
\item We identify and quantify four registration errors in the use of renderer buffers and formulate the two-lattice warp needed to correct them, within the existing warps (Section~\ref{sec:registration}).
\item We build a simple registered network with a normalised accumulator and a small temporal weight. It improves the best published PSNR on QRISP by 0.34 to 0.62\,dB at $\times$2, $\times$3 and $\times$4. At $\times$2, its pixel standard deviation on the benchmark's static segment is about one quarter that of DLSS~2, with higher ColorVideoVDP scores on every test sequence. It transfers unchanged to Arm's jittered Bistro capture~\cite{arm2025nss} and exceeds STSSNet's average PSNR when trained on one STSS scene (Section~\ref{sec:bench}). At $\times$4 on an H200, the $96\times4$ variant exceeds QRISP-L's PSNR by 0.07\,dB at 25\% lower latency for the complete recurrent step.
\end{itemize}
%% ======== END TEXT 01_introduction ========

\section{Related work}\label{sec:related}
%% ======== BEGIN TEXT 02_related ========
\paragraph*{Temporal anti-aliasing and upsampling.}
Reprojection caches reuse shading from previous frames by following motion
vectors~\cite{nehab2007accelerating,scherzer2012temporal}, and amortized supersampling accumulates jittered samples
over time with a recursive exponential filter~\cite{yang2009amortized}. Temporal anti-aliasing in games follows
the same scheme, adds history rectification such as neighbourhood clamping~\cite{karis2014high} and variance
clipping~\cite{salvi2016excursion}, and upsamples by accumulating samples at their jittered positions on the
output grid~\cite{yang2020survey,amd2022fsr2}. All of them trade lag, when stale history is kept, against flicker,
when too few samples are averaged~\cite{scherzer2012temporal,iglesias2016pixel,schied2018gradient,yang2020survey}. We show that the training loss can push a
learned accumulator to one end of this trade-off (Section~\ref{sec:mechanism}).

\paragraph*{Learned supersampling.}
Neural supersampling replaces the hand-designed accumulation with a network that sees the current samples, depth,
motion and reprojected history, e.g.~\cite{guo2022classifier}. NSRR~\cite{xiao2020neural} reconstructs from several warped frames, QRISP
networks~\cite{mercier2023efficient} run a small jitter-conditioned trunk at input resolution and
blend with the warped output, and mobile or frame-generating variants trade quality for
cost~\cite{yang2023mnss,guo2021extranet,wu2023extrass,he2024low}, on mobile even without a network at run
time~\cite{yang2024mobfgsr}. Other work adds high-resolution G-buffers~\cite{zhong2023fusesr} or demodulates
radiance~\cite{li2024neural}; industrial upscalers follow the same recipe~\cite{liu2020dlss,intel2022xess,arm2025nss},
as do joint denoisers and supersamplers for path tracing~\cite{thomas2022temporally} and minimal networks for
temporal anti-aliasing~\cite{herveau2023minimal}. We analyse the interaction between reprojection and temporal weighting.

\paragraph*{Temporal losses.}
Temporal consistency losses compare a frame with its motion-compensated predecessor. In video processing they
penalise the warped difference of the output~\cite{ruder2016artistic,lai2018learning} or match temporal statistics of the output to
those of the reference~\cite{chu2020learning}. Recurrent Monte Carlo denoisers penalise the difference between the
temporal derivatives of output and reference~\cite{chaitanya2017interactive}, and learned
supersamplers and adaptive samplers use warped or unwarped variants of
it~\cite{hasselgren2020neural,thomas2022temporally,herveau2023minimal}. In derivative-matching terms a frozen output
pays exactly the reference's own change; we show when this rewards freezing.

\paragraph*{Measuring temporal stability.}
Reference-based temporal metrics compare output and reference
changes~\cite{chu2020learning}; temporal PSNR does so after a warp and so depends on the motion field. QRISP
reports pixel standard deviation on a static segment~\cite{mercier2023efficient}, which rewards freezing or
blur. We add \protect\reflectbox{F}LIP~\cite{andersson2020flip} and
ColorVideoVDP~\cite{mantiuk2024colorvideovdp} to assess image and video quality.
%% ======== END TEXT 02_related ========

\section{Formulation}\label{sec:background}
%% ======== BEGIN TEXT 03_background ========
At frame $t$, the renderer provides jittered low-resolution colour, depth and motion. We reconstruct an output
$\hat y_t$ at scale $s\in\{2,3,4\}$ against a supersampled reference $y_t$. The changing sample positions supply
complementary information even when the scene and camera are static. Our formulation separates the accumulation
of that information from its registration between frames.

\paragraph*{Normalised accumulation.}
Following temporal upsampling~\cite{yang2009amortized,yang2020survey}, we keep a weighted colour sum $S_t$ and an
accumulated weight $N_t$ on the output lattice. The network predicts a candidate colour $\hat c_t$, a reliability
$r_t\in(0,1)$ and a retention $\rho_t\in(0,1)$ for each output pixel:
\begin{equation}
  \begin{aligned}
  S_t &= \rho_t\,v_t\,\mathcal{R}_t[S_{t-1}] + w_t\,\hat c_t,\\
  N_t &= \rho_t\,v_t\,\mathcal{R}_t[N_{t-1}] + w_t,\qquad
  \hat y_t = S_t/N_t ,
  \end{aligned}\label{eq:accum}
\end{equation}
where $\mathcal{R}_t$ reprojects the history, $v_t$ marks valid history, and $w_t=k(d_t)r_t$ weights the new
sample by its reliability and its distance $d_t$ from the pixel centre. A common rescaling of $S_t$ and $N_t$
caps the accumulated weight at $N_{\max}$ without changing the output. Before that rescaling, the new sample's
share is
\begin{equation}
  \alpha_t=w_t/N_t . \label{eq:alpha}
\end{equation}
Where the previous accumulated weight varies slowly in space, this gives
$\hat y_t\approx\alpha_t\hat c_t+(1-\alpha_t)\mathcal{R}_t[\hat y_{t-1}]$.
For valid history, the exact history estimate is $\mathcal{R}_t[S_{t-1}]/\mathcal{R}_t[N_{t-1}]$: warping the sum and weight separately need not equal warping their ratio. Thus $\alpha_t$ measures uptake of the new candidate, while the approximate blend describes how that uptake changes the output. Keeping $N_t$ lets the same predicted reliability produce a smaller update after several retained observations than after a history reset. Low retention or invalid history increases the new candidate's relative share. QRISP~\cite{mercier2023efficient} predicts a blend weight; Sections~\ref{sec:generality} and~\ref{sec:ablations} compare both forms.

For valid, spatially constant history with constant $w$ and $0<\rho<1$, the uncapped recurrence has steady-state weight $N=w/(1-\rho)$ and new-sample share $\alpha=1-\rho$. In this limit it becomes an exponential average. With changing reliability or visibility, $N_t$ records how much weighted support survived earlier frames. The retained support helps reconcile new jittered samples with previous observations; high retention also allows training to reduce their contribution. We therefore inspect deployed sample uptake alongside reconstruction accuracy when varying the temporal weight.

\paragraph*{Training objective.}
We combine per-frame reconstruction with a motion-compensated temporal derivative term:
\begin{equation}
  \begin{aligned}
  \mathcal{L}&=\sum_t\|\hat y_t-y_t\|_2^2
  +\lambda_T\sum_t\|\Delta\hat y_t-\Delta y_t\|_1,\\
  \Delta x_t&=x_t-\mathcal{R}_t[x_{t-1}] .
  \end{aligned}\label{eq:loss}
\end{equation}
The temporal term is averaged over valid pixels; training omits the first two frames of each clip.
As in recurrent denoisers and supersamplers~\cite{chaitanya2017interactive,thomas2022temporally}, it matches the
output's change to the reference's change. Both derivatives use the same reprojection, so their matching depends on registration.
%% ======== END TEXT 03_background ========

%% ======== FIGURE 2 (overview) ========
\begin{figure*}[t]
  \centering
  \includegraphics[width=\textwidth]{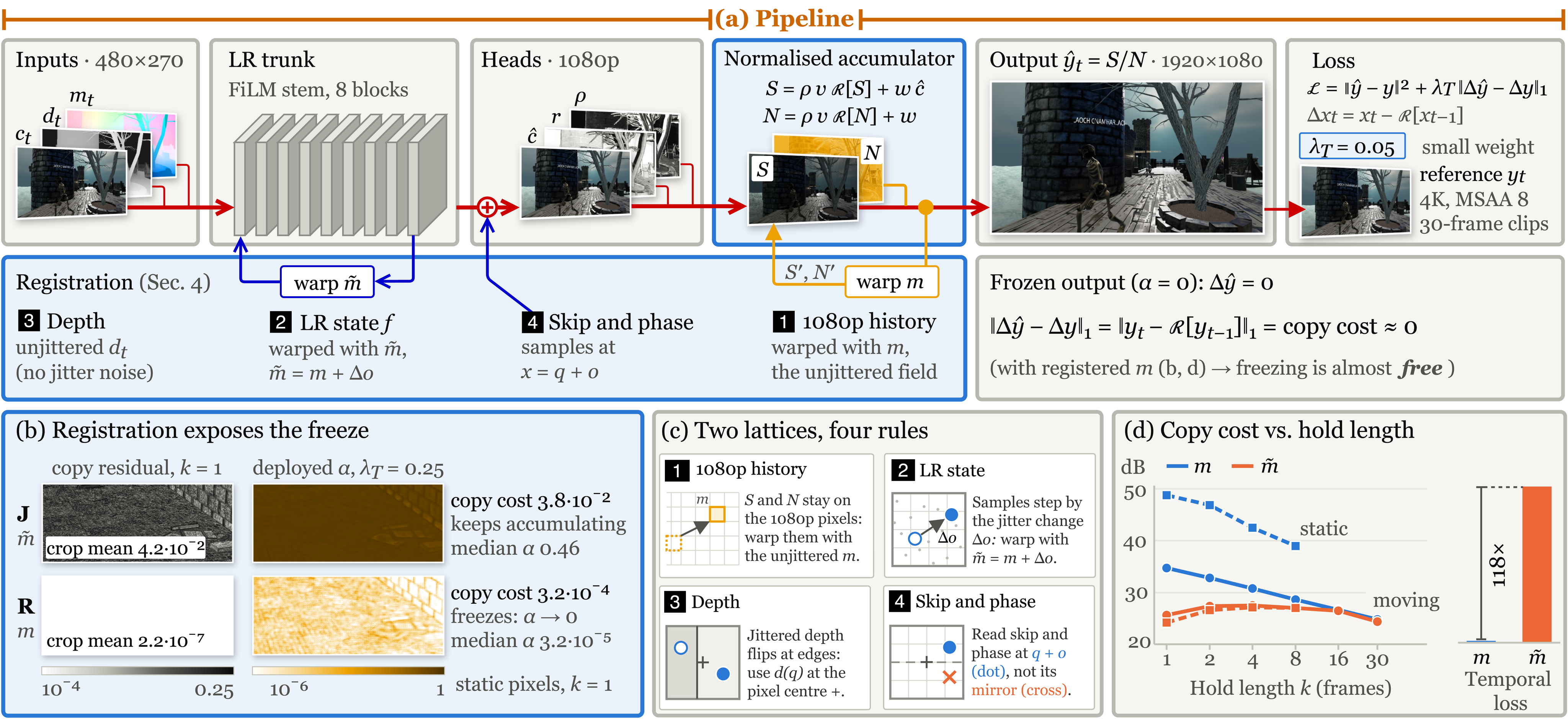}
  \caption{Registered supersampling. (a) The low-resolution trunk predicts colour, reliability and retention for the
normalised output accumulator; the frozen-output limit assumes locally constant accumulated weight. (b) A static reference has small copy cost under R; a large temporal weight then
suppresses new samples. (c) The output and sample lattices require different motion fields.
(d) Model-free copy cost over chained reprojections. Section~\ref{sec:registration} details the registration corrections; Section~\ref{sec:frozen} derives the copy cost.}
  \label{fig:pipeline}
  \vspace{-0.5em} % local float gap adjustment (<= 4.5 pt)
\end{figure*}

\section{Sample registration}\label{sec:registration}
%% ======== BEGIN TEXT 04_registration ========
The output history and the rendered samples occupy different lattices (Figure~\ref{fig:pipeline}c). The output
lattice is fixed. The input lattice moves with the camera jitter $J_t=(j_t^x,j_t^y)$: in low-resolution pixel
units, with the image $y$ axis pointing down, sample $q$ lies at
\begin{equation}
  x_t(q)=q+o_t,\qquad o_t=(j_t^x,-j_t^y). \label{eq:offset}
\end{equation}
Let $m_t$ describe surface motion between unjittered projections. We warp output-lattice quantities with $m_t$,
and sample-attached quantities, including the low-resolution recurrent features, with
\begin{equation}
  \tilde m_t=m_t+(o_t-o_{t-1}). \label{eq:lrfield}
\end{equation}
Each history buffer follows the coordinates in which it is stored. Low-resolution recurrent features move with the input sample lattice, while the output accumulator represents fixed display pixels. In a static scene, feature transport still contains the jitter difference and output transport is stationary. New samples can thus supply complementary information without displacing the accumulated image. Positions and motion must use the same units and axis convention. The following four errors violate these requirements; model-free checks quantify their effects before learning.

\paragraph*{(1) Jitter in the output-history warp.}
QRISP supplies motion between jittered projections, $\tilde m_t$. Applying it directly to the output history adds
a jitter displacement even on static content. QRISP's own warping module removes this
displacement~\cite{mercier2023efficient}, following the TAA convention of Karis~\shortcite{karis2014high}.
Our J experiments deliberately use the shipped field without compensation; QRISP's published pipeline corrects it. On static pixels, this raises the reference's one-frame copy cost by $118\times$
(Section~\ref{sec:oracle}).

\paragraph*{(2) One field for both lattices.}
Removing jitter from every warp creates a different error: $m_t$ omits the displacement of the input lattice.
Carrying the previous low-resolution render with $\tilde m_t$ reaches 28.35\,dB on the test set, compared with
26.52\,dB using $m_t$. The mean absolute residual rises from 0.0357 to 0.0433, or 21\%.
The correction therefore depends on where the state is stored.

\paragraph*{(3) Jittered depth in history tests.}
Jittered depth samples different surface points each frame. Depth changes can then trigger disocclusion
tests~\cite{schied2017spatiotemporal} or motion dilation even when the camera is stationary.
On the School stop, the mean absolute reprojected depth change is about $10^3$ times larger for jittered depth.
We use the unjittered depth supplied by the renderer.

\paragraph*{(4) Incorrect sample positions in the skip and phase.}
A bicubic skip must place the current samples at $x_t(q)$, rather than at the input pixel centres. Correcting this
offset gains 0.86 to 1.53\,dB on five moving SeaPort frames without a network. Likewise, using $+j_t^y$ in the
sample phase mirrors its vertical offset. At $\times$4 this selects the wrong row for all 16 jitter phases,
with a mean displacement of 1.80 output pixels. On the School stop, a plain splat reaches 34.19\,dB with the
correct phase, 29.78\,dB with the mirrored phase, and 28.38\,dB with a bicubic upsampling of the jittered input.
Supplementary Table~\ref{supp:tab:modelfree} collects these diagnostics.

Training can conceal the errors. At $\lambda_T=0$, our network reaches 29.72\,dB with the original conventions
and 29.75\,dB with full registration. Individual component removals also produce small PSNR changes
(Table~\ref{tab:ablation}c). Accuracy alone reveals little of these errors; their effect on the temporal-loss response is much larger.
%% ======== END TEXT 04_registration ========

\section{How temporal loss suppresses accumulation}\label{sec:mechanism}
%% ======== BEGIN TEXT 05_mechanism ========
A frozen output pays a copy cost determined by the reference and its registration.
We measure this cost on reference images and test its predicted effect on sample uptake in trained networks. The oracle isolates the motion field.
The training experiments compare the complete registration conventions of Section~\ref{sec:registration}.

\subsection{Loss of a frozen output}\label{sec:frozen}
Define a frozen output by $\hat y_t=\mathcal{R}_t[\hat y_{t-1}]$, so $\Delta\hat y_t=0$. Its temporal penalty equals the reference's reprojected change:
\begin{equation}
  \|\Delta\hat y_t-\Delta y_t\|_1
  =\|y_t-\mathcal{R}_t[y_{t-1}]\|_1 . \label{eq:frozen}
\end{equation}
We call the reference's reprojected change the \emph{copy cost}. Writing $e_t=\hat y_t-y_t$, the frozen
reconstruction error satisfies
\[
  \|e_t\|\leq\|\mathcal{R}_t[e_{t-1}]\|
  +\|\mathcal{R}_t[y_{t-1}]-y_t\| .
\]
If the reference is unchanged after reprojection, copying minimises the temporal term regardless of the retained error. A small nonzero copy cost keeps the temporal penalty small, while the reconstruction term continues to penalise retained error. On registered static content, a sufficiently large $\lambda_T$ can favour keeping this imperfect history over incorporating fluctuating new estimates.

Jitter-carrying motion raises the copy cost by moving an otherwise static reference. The derivation applies to
derivative-matching losses used in recurrent denoising and supersampling~\cite{chaitanya2017interactive,thomas2022temporally,herveau2023minimal};
whether optimisation actually freezes depends on the loss norm, weight and reconstruction term.
Losses on the output's own change~\cite{lai2018learning} use a different target.

%% ======== FIGURE 3 (F_dial4) ========
\begin{figure*}[t]
  \centering
  \includegraphics[width=\textwidth]{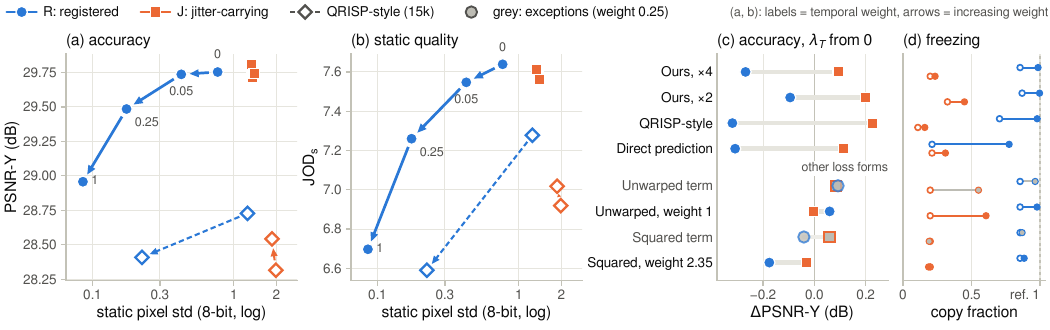}
  \caption{Registration and temporal weight. (a,b) Test-set PSNR and static JOD against static pixel STD for
  $\lambda_T\in\{0,0.05,0.25,1\}$; arrows indicate increasing weight. Filled markers: our $\times$4 network,
  45k iterations; hollow diamonds: QRISP-style network, 15k. (c,d) PSNR change and static copy fraction when the
  weight rises from 0 to 0.25, across architectures, scales and loss forms. R uses full registration; J uses the
  original misregistered conventions. Under R, the warped $\ell_1$ term increases copying while reducing
  accuracy and static JOD. Grey rows test unwarped and squared terms, including larger weights.
  Section~\ref{sec:generality} defines copy fraction; Supplementary Tables~\ref{supp:tab:grid}--\ref{supp:tab:split} give full results.}
  \label{fig:dial}
  \vspace{-0.5em} % local float gap adjustment (<= 4.5 pt)
\end{figure*}

\subsection{Copy cost without a network}\label{sec:oracle}
We carry the QRISP reference $y_{t-k}$ forward through $k$ chained reprojections and compare it with $y_t$,
using either $m_t$ or the shipped jitter-carrying field. All comparisons use pixels valid for every hold length
and both fields (Figure~\ref{fig:pipeline}d).

At $\times$4, the one-frame registered copy reaches 35.71\,dB over the test frames, compared with 25.45\,dB for
the jitter-carrying warp. On static-camera frames it reaches 48.74\,dB, while the misregistered copy stays at
24 to 27\,dB. The registered copy exceeds 58\,dB through four reprojections on the longest stop.
The corresponding static $\ell_1$ copy costs are $3.2\times10^{-4}$ and $3.76\times10^{-2}$, a $118\times$
ratio; it rises to $1300\times$ on the longest stop and falls to $2.2\times$ on moving frames.
The one-frame PSNR gap also persists at $\times$3 and $\times$2 (Supplementary Table~\ref{supp:tab:modelfree}).
Thus the motion convention alone strongly changes the penalty for copying static history.

\subsection{Response to temporal weight}\label{sec:dial}
We train the same 192-channel, eight-block network at $\times$4 for 45k iterations with
$\lambda_T\in\{0,0.05,0.25,1\}$, holding data, schedule and seed fixed. R applies all four registration fixes.
J uses jitter-carrying motion on both lattices, jittered depth, an unregistered skip and a mirrored phase.
These runs compare complete registration pipelines; the model-free oracle isolates the motion field.

Under R, raising $\lambda_T$ from 0 to 1 reduces School-stop pixel STD from 0.77 to 0.086 grey levels and deployed
median $\alpha$ from $3.4\times10^{-2}$ to $9.6\times10^{-7}$. The output takes in almost no new information.
Accuracy and perceived quality fall: test-set PSNR decreases from 29.75 to 28.96\,dB and mean static JOD from
7.64 to 6.70 (Figure~\ref{fig:dial}a,b). At $\lambda_T=0.25$, the PSNR loss is 0.86\,dB on the 97 static-camera
frames and 0.24\,dB on the 1363 moving frames (Table~\ref{tab:flip}). The per-frame penalty is about 3.6 times greater on static views, where unchanged correspondence makes copying cheap. Temporal PSNR nevertheless rises from
30.59 to 31.32\,dB across the sweep. Temporal PSNR rewards the increasingly frozen output as accuracy and static JOD fall.

Under J, static STD remains between 1.33 and 1.40 and median $\alpha$ between 0.39 and 0.61.
The loss barely changes static variation, while test-set PSNR rises by at most 0.09\,dB.
Training retention statistics are reported in Supplementary Section~\ref{supp:section-4}.

\subsection{Architecture, scale and loss form}\label{sec:generality}
%% ======== TABLE 1 (T3_flip) ========
\begin{table}[t]
\centering
\caption{PSNR-Y change from $\lambda_T=0$ to 0.25 (last row: 0.05), relative to the same network and pipeline.
The warped $\ell_1$ term costs more accuracy on static than moving frames under R.}
\label{tab:flip}
\fontsize{7.4}{8.7}\selectfont
\setlength{\tabcolsep}{2.1pt}
\begin{tabular*}{\columnwidth}{@{\extracolsep{\fill}}l c c c c c c@{}}
\toprule
 & \multicolumn{3}{c}{J: $\Delta$PSNR-Y} & \multicolumn{3}{c}{R: $\Delta$PSNR-Y} \\
\cmidrule(lr){2-4}\cmidrule(l){5-7}
Setting & static & moving & all & static & moving & all \\
\midrule
Ours, $\times$4 & +0.11 & +0.10 & +0.09 & $-$0.86 & $-$0.24 & $-$0.27 \\
Ours, $\times$2 & +0.23 & +0.20 & +0.20 & $-$0.68 & $-$0.05 & $-$0.09 \\
QRISP-style$^{a}$ & +0.31 & +0.23 & +0.23 & $-$0.85 & $-$0.30 & $-$0.32 \\
Direct prediction & +0.12 & +0.12 & +0.11 & $-$0.70 & $-$0.29 & $-$0.31 \\
\midrule
Ours, $\lambda_T{=}0.05$ & +0.05 & +0.07 & +0.07 & $-$0.19 & 0.00 & $-$0.02 \\
\bottomrule
\end{tabular*}
\par\smallskip
\begin{minipage}{\linewidth}\fontsize{6.8}{8}\selectfont
$\times$4, 45k iterations unless noted; $^{a}$15k. Static/moving: frame means over 97/1363 frames;
all: mean over the three scenes. Supplementary Table~\ref{supp:tab:split} includes the other loss forms.
\end{minipage}
\end{table}

We repeat the $\lambda_T=0$ versus 0.25 comparison at $\times$2 and with two alternative networks at
$\times$4: a QRISP-style direct-blend architecture~\cite{mercier2023efficient}, and a direct-prediction network
with reprojected colour history but no accumulator or blend weight. The QRISP-style models use our R and J pipelines and 15k training iterations. Published QRISP uses jitter-corrected reprojection (Section~\ref{sec:registration}).

For architectures without an explicit $\alpha$, we measure the \emph{copy fraction}: the share of in-bounds static
pixels whose output differs from its reprojected previous output by less than $0.5/255$ in every RGB channel.
Static pixels are those where the reference meets the same threshold, so the reference scores 1.
High copy fraction can reflect correct convergence; we identify harmful freezing when increased copying
coincides with reduced accuracy.

Across all three network families and both scales, the warped $\ell_1$ term lowers PSNR under R and raises it
slightly under J (Table~\ref{tab:flip}). Under R, static JOD drops by 0.14 to 0.69 and copy fraction reaches
0.77 to 1.00; under J it remains below 0.5 (Figure~\ref{fig:dial}c,d).
Freezing also appears in the direct-prediction network. Replacing its colour history by
the current registered bicubic image at each frame raises static STD from 0.55 to 2.79; resetting only recurrent
features raises it to 0.70 (Supplementary Section~\ref{supp:section-5}). Its low variation depends mainly on colour-history reuse.

Loss form changes the response. The unwarped derivative $\Delta x_t=x_t-x_{t-1}$, used
in~\cite{chaitanya2017interactive,hasselgren2020neural}, promotes copying under both conventions.
At $\lambda_T=1$ it gives an R copy fraction of 0.98 and median $\alpha=4\times10^{-5}$.
It lowers static-frame PSNR and JOD, but gains moving-frame PSNR, leaving the aggregate PSNR nearly unchanged.
The squared derivative term at $\lambda_T=0.25$ has little effect. At 2.35, chosen to match the median logged
loss value of the $\ell_1$ term over the second half of training, it costs 0.18\,dB under R but leaves median
$\alpha$ near 0.03 and copy fraction at 0.88. The weights align logged loss values, while the two norms apply different gradients. For nonzero residuals, the $\ell_1$ term has a constant-magnitude subgradient; the squared term scales its gradient with residual size.

On unjittered STSS~\cite{he2024low}, $\lambda_T=0.25$ reduces all-frame Arena PSNR from 34.16 to 30.46\,dB (Supplementary Table~\ref{supp:tab:second}).

\subsection{Implications for training}\label{sec:consequences}
We use $\lambda_T=0.05$ on QRISP. In the 45k analysis runs it reduces static STD from 0.77 to 0.43 at $\times$4
and from 0.41 to 0.10 at $\times$2, with no moving-frame PSNR loss, about 0.2\,dB lower static-frame PSNR, and
0.04 to 0.09 lower static JOD. At $\times$4, median static $\alpha=2.3\times10^{-3}$ indicates a mild suppression
of sample uptake. We choose this weight for the measured trade-off between accuracy and temporal variation.
It must be reconsidered after changing registration conventions or the data distribution.

Normalised accumulation supplies an additional reduction in static variation at comparable PSNR
(Table~\ref{tab:ablation}a). We evaluate both choices with reconstruction and perceptual metrics alongside
STD and temporal PSNR. Stability metrics alone would favour the larger weights that freeze the output and reduce accuracy. To diagnose freezing, a static test should keep sampling jitter active and report accuracy together with sample uptake. Turning jitter off removes the complementary samples needed to test whether the reconstruction continues to use new evidence.
%% ======== END TEXT 05_mechanism ========

\section{Model and training}\label{sec:model}
%% ======== BEGIN TEXT 06_model ========
Our recurrent model couples a low-resolution trunk to the registered output accumulator (Figure~\ref{fig:pipeline}a).

\paragraph*{Network.}
At input resolution, the trunk receives jittered colour, unjittered depth and motion, a disocclusion mask, eight
neighbouring depth differences, the output history and accumulated weight rearranged by space-to-depth,
reprojected recurrent features, and sample phase. The first $3\times3$ convolution is modulated by jitter,
following jitter-conditioned reconstruction~\cite{mercier2023efficient,herveau2023minimal}.
Each residual block has two $3\times3$ convolutions with GELU, no normalisation, and a learned channel scale
initialised to 0.1. Final models use eight blocks, with 192 channels at $\times$2 and $\times$3 and 256 at
$\times$4; mechanism runs use 192 channels.

Three heads predict $\hat c_t$, $r_t$ and $\rho_t$ on the output lattice through sub-pixel rearrangement.
The colour head adds a residual to a bicubic skip evaluated at the registered sample positions.
The accumulator uses $N_{\max}=16$ and a Gaussian distance kernel whose learned width approaches a nearly flat
profile. Normalised weighted accumulation is established in temporal upsampling~\cite{yang2009amortized,yang2020survey,amd2022fsr2};
here the network learns reliability and retention. A state guard resets pixel blocks whose colour or recurrent
features leave a plausible range, protecting long sequences against rare divergence. History validity $v_t$ gates both the sum and weight. A disocclusion clears accumulated support at the affected pixel, making the output equal to the current candidate. Elsewhere, the recurrence retains evidence from earlier jitter phases. It thus combines accumulation in a settled view with replacement of history when visibility changes.

\paragraph*{Training.}
QRISP training uses 30-frame clips, $66\times66$ input-resolution crops and batch size 8. Each clip shares a crop
and random horizontal and vertical flips, applied consistently to colour, motion and jitter.
Equation~\ref{eq:loss} uses 8-bit sRGB scaled to $[0,1]$ and $\lambda_T=0.05$, omitting the first two clip frames.
We use AdamW~\cite{loshchilov2019decoupled} ($\beta_1=0.9$, $\beta_2=0.99$, weight decay $10^{-4}$) with a
one-cycle schedule~\cite{smith2019super} and peak learning rate $6\times10^{-4}$.
Analysis models use 45k iterations; final models use 100k at $\times$2 and $\times$3 and 200k at
$\times$4. The 15k registration controls use 16-frame clips without temporal loss; STSS training uses 16-frame clips.

\paragraph*{Deployment and smaller variants.}
The trunk runs in bfloat16 with float32 accumulation. Fused Triton kernels~\cite{tillet2019triton} implement input
preparation, warps, convolutions and accumulation within one CUDA graph.
The fused float32 step differs from the scoring implementation by at most 0.0003\,dB; the timed bfloat16 step
differs by less than 0.01\,dB.
For the latency study, 64- and 96-channel trunks use re-parameterised
blocks~\cite{ding2021repvgg,zhang2021edge} folded into plain convolutions.
Shift-MLP variants fuse spatial channel shifts with a two-layer MLP and use a peak rate of $2\times10^{-3}$.
Supplementary Sections~\ref{supp:section-11}--\ref{supp:section-13} detail architectures and timings. Code, models and the registered pipeline
will be released. Code page: \url{https://github.com/CV-AC/Frozen_history}.
%% ======== END TEXT 06_model ========
%% ======== BEGIN TABLE 2 (T1_headline) ========

\begin{table*}[!t]
\centering
\caption{QRISP quality and run time. $^\dagger$Published QRISP/NSRR scores; other metrics use our harness.
Static PSNR and STD use the 18-frame School stop; static FLIP uses 97 static-camera frames;
JOD$_{\mathrm s}$ averages three stops. ms: one recurrent $\times$4 step on an H200, excluding rendering.
Metrics and evaluation populations are defined in Section~\ref{sec:setup}. Dashes indicate unavailable metrics or timings.}
\label{tab:headline}
\fontsize{7.5}{9}\selectfont
\setlength{\tabcolsep}{2.8pt}
\begin{tabular*}{\textwidth}{@{\extracolsep{\fill}}l c c c c c c c c c c c c@{}}
\toprule
Method & Iter. & ms$\,\downarrow$ & PSNR-Y$\,\uparrow$ & SSIM-Y$\,\uparrow$ & LPIPS$\,\downarrow$ & FLIP$\,\downarrow$ & JOD$\,\uparrow$ & tPSNR$_{60}\uparrow$ & Static PSNR$\,\uparrow$ & Static FLIP$\,\downarrow$ & Static std & JOD$_{\mathrm s}$ \\
\midrule
\multicolumn{13}{@{}l}{\textit{$\times$2 (960$\times$540 $\rightarrow$ 1920$\times$1080)}} \\
Ours (192$\times$8) & 100k & --- & \textbf{32.69} & \textbf{0.9246} & \textbf{0.091} & \textbf{0.0570} & \textbf{8.61} & 33.57 & \textbf{37.57} & \textbf{0.0458} & 0.096 & \textbf{8.26} \\
DLSS 2 (shipped frames) & --- & --- & 30.12 & 0.8821 & 0.151 & 0.0732 & 8.22 & 31.55 & 36.25 & 0.0546 & 0.41 & 7.96 \\
TAAU (classical) & --- & --- & 29.99 & 0.8740 & 0.162 & 0.0743 & 7.99 & 31.09 & 35.43 & 0.0568 & 0.67 & 7.61 \\
FSR 1.2~\cite{amd2021fsr1} (shipped, spatial)$^{\P}$ & --- & --- & 29.44 & 0.8668 & 0.160 & 0.0708 & 7.88 & 28.49 & 32.80 & 0.0585 & 0.010 & 7.08 \\
Bicubic (our harness) & --- & --- & 29.31 & 0.8649 & 0.217 & 0.0719 & 7.81 & 28.40 & 32.63 & 0.0593 & 0.011 & 7.01 \\
QRISP-L~\cite{mercier2023efficient}$^{\dagger}$ & 500k & --- & 32.21 & 0.9115 & --- & --- & --- & --- & --- & --- & --- & --- \\
QRISP-M~\cite{mercier2023efficient}$^{\dagger}$ & 500k & --- & 31.80 & 0.9044 & --- & --- & --- & --- & 37.38 & --- & 0.54 & --- \\
QRISP-S~\cite{mercier2023efficient}$^{\dagger}$ & 500k & --- & 31.18 & 0.8941 & --- & --- & --- & --- & --- & --- & --- & --- \\
NSRR~\cite{xiao2020neural}$^{\dagger}$ & --- & --- & 31.89 & 0.9075 & --- & --- & --- & --- & --- & --- & --- & --- \\
\midrule
\multicolumn{13}{@{}l}{\textit{$\times$3 (640$\times$360 $\rightarrow$ 1920$\times$1080)}} \\
Ours (192$\times$8) & 100k & --- & \textbf{31.01} & \textbf{0.8899} & \textbf{0.140} & \textbf{0.0697} & \textbf{8.14} & 32.09 & \textbf{36.00} & \textbf{0.0542} & 0.25 & \textbf{7.95} \\
TAAU (classical) & --- & --- & 28.59 & 0.8299 & 0.233 & 0.0876 & 7.36 & 29.91 & 33.40 & 0.0674 & 0.86 & 7.02 \\
Bicubic (our harness) & --- & --- & 27.37 & 0.8009 & 0.370 & 0.0999 & 6.83 & 26.55 & 30.50 & 0.0833 & 0.0090 & 6.03 \\
QRISP-L~\cite{mercier2023efficient}$^{\dagger}$ & 500k & --- & 30.67 & 0.8747 & --- & --- & --- & --- & --- & --- & --- & --- \\
QRISP-M~\cite{mercier2023efficient}$^{\dagger}$ & 500k & --- & 30.23 & 0.8655 & --- & --- & --- & --- & --- & --- & --- & --- \\
NSRR~\cite{xiao2020neural}$^{\dagger}$ & --- & --- & 30.24 & 0.8729 & --- & --- & --- & --- & --- & --- & --- & --- \\
\midrule
\multicolumn{13}{@{}l}{\textit{$\times$4 (480$\times$270 $\rightarrow$ 1920$\times$1080)}} \\
Ours (256$\times$8) & 200k & 2.032 & \textbf{30.04} & \textbf{0.8635} & \textbf{0.175} & \textbf{0.0801} & \textbf{7.77} & 31.33 & \textbf{34.85} & \textbf{0.0612} & 0.38 & \textbf{7.63} \\
Ours (192$\times$8) & 200k & 1.258 & 29.92 & 0.8615 & 0.179 & 0.0806 & 7.75 & 31.16 & 34.75 & 0.0619 & 0.41 & 7.61 \\
Ours (96$\times$8)$^{r}$ & 200k & 0.790 & 29.76 & 0.8558 & 0.190 & --- & --- & 30.90 & 34.61 & --- & 0.46 & --- \\
Ours (96$\times$4)$^{r}$ & 200k & 0.524 & 29.49 & 0.8457 & 0.208 & --- & --- & 30.68 & 34.57 & --- & 0.53 & --- \\
Ours (64$\times$4)$^{r}$ & 200k & 0.332 & 29.32 & 0.8394 & 0.218 & --- & --- & 30.54 & 34.46 & --- & 0.46 & --- \\
Ours, shift-MLP (64$\times$8) & 200k & 0.301 & 29.35 & 0.8405 & 0.211 & --- & --- & 30.58 & 34.54 & --- & 0.45 & --- \\
Ours, shift-MLP (64$\times$6) & 200k & 0.201 & 29.32 & 0.8391 & 0.213 & --- & --- & 30.49 & 34.51 & --- & 0.51 & --- \\
TAAU (classical) & --- & --- & 27.61 & 0.7938 & 0.295 & 0.0997 & 6.82 & 29.08 & 31.92 & 0.0774 & 1.06 & 6.46 \\
Bicubic (our harness) & --- & --- & 26.15 & 0.7534 & 0.444 & 0.1225 & 6.11 & 25.40 & 29.35 & 0.1034 & 0.0090 & 5.26 \\
QRISP-L~\cite{mercier2023efficient}$^{\dagger}$ & 500k & 0.698$^{\ast}$ & 29.42 & 0.8403 & --- & --- & --- & --- & --- & --- & --- & --- \\
QRISP-M~\cite{mercier2023efficient}$^{\dagger}$ & 500k & 0.443$^{\ast}$ & 29.06 & 0.8305 & --- & --- & --- & --- & --- & --- & --- & --- \\
NSRR~\cite{xiao2020neural}$^{\dagger}$ & --- & --- & 29.02 & 0.8364 & --- & --- & --- & --- & --- & --- & --- & --- \\
\bottomrule
\end{tabular*}
\par\smallskip
\begin{minipage}{\linewidth}\scriptsize
$^\dagger$Mercier et al.~\cite{mercier2023efficient}; 500k iterations for QRISP. $^\P$Spatial FSR, supplied at
$\times$2 only. Bicubic/FSR use unjittered inputs; learned models and TAAU use jittered inputs.
$^\ast$Published architecture re-implemented with random weights, paired with published PSNR.
$^r$Re-parameterised blocks folded for inference. Timings: batch 1, fused kernels, one CUDA graph, bf16/fp32;
300 calls after 50 warm-ups. Static STD: 8-bit output, School frames 275--292; reference 0.010.
STD and tPSNR are not ranked because both reward freezing; bold marks the best other scores per scale.
\end{minipage}
\vspace{-0.5em} % local float gap adjustment (< 4 pt at the table font)
\end{table*}

\section{Experiments}\label{sec:experiments}
%% ======== BEGIN TEXT 07_experiments ========
Section~\ref{sec:mechanism} tests the freezing mechanism. Here we assess reconstruction quality, practical
temporal behaviour, component contributions, transfer and computation cost.
%% ======== END TEXT 07_experiments ========

%% ======== FIGURE 4 (Q9_results) ========
\begin{figure*}[t]
  \centering
  \includegraphics[width=\textwidth]{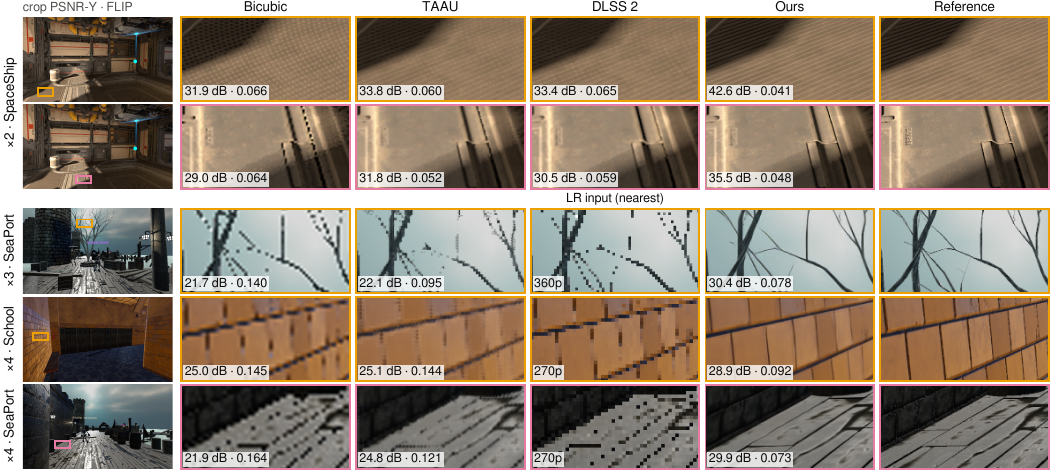}
  \caption{Reconstructions at $\times$2, $\times$3 and $\times$4, with crop PSNR-Y and \reflectbox{F}LIP.
Crops maximise our PSNR gain within detailed regions ($\times$2, $\times$4 School: luma $\geq$ 100), against DLSS~2
at $\times$2, the best baseline at $\times$3 and $\times$4 School, and bicubic at $\times$4 SeaPort.
Table~\ref{tab:headline} reports test-set scores.
At $\times$3 and $\times$4 the third column is the nearest-neighbour LR input because no DLSS frames are supplied.}
  \label{fig:x2}
\end{figure*}

\subsection{Setup and protocol}\label{sec:setup}

%% ======== BEGIN TEXT 07a_setup ========
\paragraph*{Data and scoring.}
QRISP~\cite{mercier2023efficient} provides jittered, mip-biased inputs at 540p, 360p and 270p, motion with and
without jitter, depth, and supersampled 1080p references. Its test set contains five 300-frame camera sequences
across three scenes. We reset history per sequence and omit eight warm-up frames, leaving 1460 scored frames.
PSNR-Y and SSIM-Y are averaged over frames within each scene, then over scenes; LPIPS uses
RGB~\cite{zhang2018unreasonable}. Section~\ref{sec:bench} gives the STSS and Arm evaluation protocols.

\paragraph*{Temporal and perceptual metrics.}
Temporal PSNR compares warped output and reference derivatives using one shared unjittered motion field for all
methods, including DLSS~2; each frame-pair score is capped at 60\,dB before averaging.
Static STD is the mean per-pixel temporal standard deviation on AbandonedSchool/0001 frames 275--292, in
8-bit intensity units. Analysis uses float output; Table~\ref{tab:headline} and history ablations use 8-bit
output. The two versions differ by at most about 0.05.
Neither STD nor temporal PSNR distinguishes successful reconstruction from freezing.
We therefore also report \reflectbox{F}LIP (LDR, 67 pixels per degree)~\cite{andersson2020flip} and
ColorVideoVDP~\cite{mantiuk2024colorvideovdp} v0.5.7, configured for a 24-inch 1080p sRGB display at
200\,cd/m$^2$ and 0.6\,m viewing distance.
JOD is computed per sequence; JOD$_{\mathrm s}$ averages three stops (School 274--292, SeaPort 20--38 and
205--223). Static FLIP and the PSNR splits of Table~\ref{tab:flip} use all 97 static-camera frames,
identified by median motion below 0.02 pixels in the unjittered 270p field.
Static PSNR and STD instead use the 18-frame School stop.

\paragraph*{Comparisons and repeatability.}
QRISP and NSRR use the published scores of Mercier et al.~\shortcite{mercier2023efficient}, marked $\dagger$. DLSS~2 v2.2 is supplied as captured frames at $\times$2 only; our harness
reproduces its published PSNR within 0.1\,dB (30.12 versus 30.21).
Our classical TAAU receives the registered inputs, splats samples with a Gaussian fit to
Blackman--Harris~\cite{karis2014high}, reprojects with Catmull--Rom, uses YCoCg variance
clipping~\cite{salvi2016excursion} and decaying accumulation weights~\cite{amd2022fsr2}.
Its parameters and those of a neighbourhood-clamp variant~\cite{karis2014high} are selected on 30 training
sequences. Bicubic and the supplied spatial FSR~1.2~\cite{amd2021fsr1} use unjittered renders.
The mechanism grid uses one seed per setting. Seven 15k seed pairs differ by 0.003 to 0.08\,dB (median 0.03; Supplementary Table~\ref{supp:tab:seeds}).
%% ======== END TEXT 07a_setup ========

\subsection{Main results}\label{sec:main}

%% ======== FIGURE 5 (Q6c_temporal) ========
\begin{figure*}[t]
  \centering
  \includegraphics[width=\textwidth]{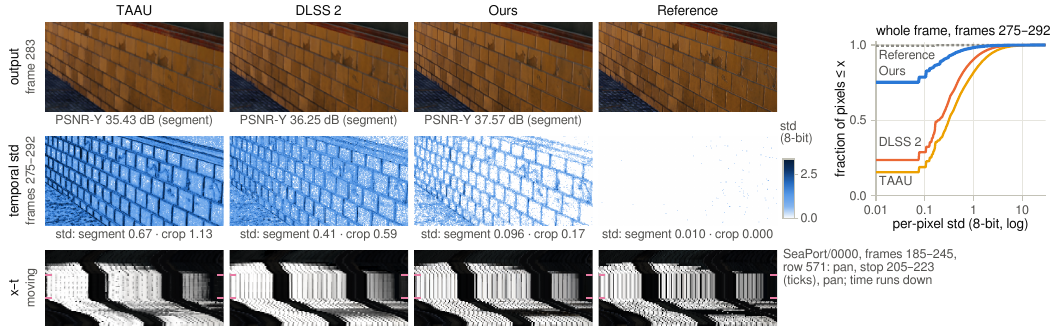}
  \caption{Temporal behaviour at $\times$2. Top: School frame 283 and static STD maps over frames 275--292;
right: full-frame STD distributions. Our mean STD is 0.096 versus DLSS~2's 0.41, with static PSNR
37.57 versus 36.25\,dB. Our 90th-percentile STD is 0.24 versus 0.97; 75\% of pixels are constant, versus
24\% for DLSS~2 and 99.5\% for the reference. Bottom: a SeaPort x--t slice across a pan--stop--pan interval.
The plotted trajectories show no visible lag; Figure~\ref{fig:evidence}c gives per-frame scores.}
  \label{fig:temporal}
  \vspace{-0.5em} % local float gap adjustment (<= 4.5 pt)
\end{figure*}

%% ======== FIGURE 6 (Q10_evidence) ========
\begin{figure}[t]
  \centering
  \includegraphics[width=\columnwidth]{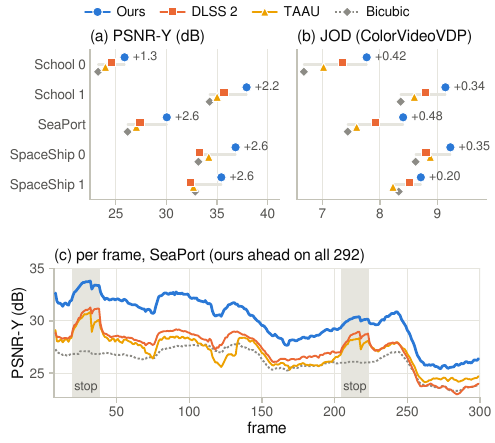}
  \caption{Whole-sequence and per-frame evidence at $\times$2. (a) PSNR-Y and (b) JOD on each test sequence;
labels give our margin over the strongest plotted baseline. (c) SeaPort PSNR-Y, with camera stops shaded.
Our model leads in PSNR on every SeaPort frame and on 1455 of the 1460 scored QRISP frames.
School and SpaceShip abbreviate AbandonedSchool and SpaceShipDemo.}
  \label{fig:evidence}
  \vspace{-0.5em} % local float gap adjustment (<= 4.5 pt)
\end{figure}

%% ======== BEGIN TEXT 07b_main ========
\paragraph*{Reconstruction quality.}
Table~\ref{tab:headline} places our final models 0.48, 0.34 and 0.62\,dB above published QRISP-L PSNR at
$\times$2, $\times$3 and $\times$4, respectively, with higher SSIM at every scale.
These benchmark scores compare complete systems; Sections~\ref{sec:mechanism} and~\ref{sec:ablations} isolate registration, temporal weighting and accumulation through controlled runs.
Against our TAAU with the same registered inputs, the learned models gain 2.4 to 2.7\,dB.
Changing classical variance clipping to neighbourhood clamping changes PSNR by at most 0.06\,dB
(Supplementary Table~\ref{supp:tab:extra-rows}).
Figure~\ref{fig:x2} illustrates reconstruction of thin and periodic structure. The crops maximise improvement within eligible detailed regions; Table~\ref{tab:headline} and Figure~\ref{fig:evidence} give aggregate and per-sequence whole-frame scores.

\paragraph*{Temporal quality at $\times$2.}
Against the captured DLSS~2 frames, our final model gains 2.57\,dB PSNR and 0.39 JOD over whole sequences.
It has higher PSNR and JOD and lower FLIP on each of the five sequences
(Figure~\ref{fig:evidence}; Supplementary Table~\ref{supp:tab:x2detail}).
On the 18-frame School stop, static STD is 0.096 versus 0.41, with static PSNR 37.57 versus
36.25\,dB (Figure~\ref{fig:temporal}). The 90th-percentile STD is 0.24 versus 0.97, and 75\% of our pixels are constant at 8-bit precision, versus 24\% for DLSS~2 and 99.5\% for the reference. Both statistics move toward the reference distribution as static accuracy improves.
Mean JOD across the three stops is 8.26 versus 7.96.
On the two SeaPort stops, where the reference itself changes, our STD is 1.05 and 1.49, compared with DLSS~2's
1.53 and 1.76. The reduction spans all three evaluated camera stops.

The controlled 45k runs isolate the temporal weight: at $\times$2, raising $\lambda_T$ from 0 to 0.05 reduces float-output School STD from 0.41 to 0.10 (Section~\ref{sec:consequences}). The zero-weight model already reaches static JOD 8.27, above DLSS~2; the final 100k model scores 8.26. The weight primarily reduces variation, while higher static JOD is already present without it. Bicubic and spatial FSR have still lower STD (0.011 and 0.010) but lower static JOD
(7.01 and 7.08), reinforcing the need to interpret variation beside reconstruction quality.

Figure~\ref{fig:temporal} shows no visible lag in the plotted SeaPort pan--stop--pan slice.
The per-frame comparison remains ahead of all plotted baselines immediately after both stop-to-pan transitions
(Figure~\ref{fig:evidence}c), and on 1455 of 1460 scored test frames in PSNR.
The reconstruction gains persist as the camera resumes motion.
%% ======== END TEXT 07b_main ========

\subsection{Additional evaluations}\label{sec:bench}
%% ======== BEGIN TEXT 07g_bench ========
\paragraph*{Zero-shot transfer to a jittered capture.}
Table~\ref{tab:bench}a evaluates the QRISP-trained $\times$2 model unchanged on Arm's Bistro
capture~\cite{arm2025nss}: one 100-frame Unreal Engine sequence at 30\,Hz, with 92 scored frames, period-32
jitter on a 1/8-pixel grid, and a 16-spp reference.
We convert motion, depth and jitter to Section~\ref{sec:registration}'s conventions and tone-map input and
reference with Arm's evaluation code. Model-free checks verify the conversions (Supplementary Table~\ref{supp:tab:arm-checks}).
Classical methods keep their QRISP parameters.
The 100k model reaches 31.62\,dB and leads the two classical temporal upsamplers by 4.4 to 4.5\,dB,
with better scores in all reported metrics. The matched 45k rows isolate the effect of retaining history:
31.07 versus 26.64\,dB, a 4.43\,dB difference.
The gain transfers across renderer and jitter conventions.

\paragraph*{Retraining without jitter.}
For STSS~\cite{he2024low}, we retrain on the first Lewis training file (2798 frames), with 16-frame clips, and
use the authors' metric code for Table~\ref{tab:bench}b.
At $\lambda_T=0$, our mean PSNR is 33.48\,dB versus STSSNet's 32.89.
The scene differences are uneven: +0.54\,dB on Arena, unseen by both methods, and +2.47\,dB on Lewis,
but $-0.60$ and $-0.04$\,dB on SunTemple and Subway, seen during training only by STSSNet.
The Lewis gain therefore contributes substantially to the mean.
At 0.05, Arena falls to 33.11\,dB, below STSSNet and bicubic; at 0.25, PSNR falls on all four scenes and
the mean drops by 1.93\,dB from the zero-weight model.
Both weights selected for the QRISP comparisons reduce accuracy here, most strongly on Arena.
At 0.25, Lewis loses 0.40\,dB, while the three unseen scenes lose 1.72 to 3.64\,dB. The temporal weight has its largest accuracy cost outside the training scene.
All-frame scores and perceptual metrics are in Supplementary Table~\ref{supp:tab:stss-ours}; their protocol differs from the
authors' even-frame, 8-bit evaluation used here.
%% ======== END TEXT 07g_bench ========

%% ======== BEGIN TABLE 3 (T7_bench) ========
\begin{table}[!t]
\centering
\caption{Other benchmarks at $\times$2. (a) Arm Bistro, zero-shot from QRISP, with matched 45k history controls.
(b) STSS, retrained on one Lewis file and scored with the authors' code. Temporal weight reduces PSNR on all
four STSS scenes.}
\label{tab:bench}
\fontsize{7.4}{8.7}\selectfont
\setlength{\tabcolsep}{2.0pt}
\begin{tabular}{@{}l c c c c c@{}}
\toprule
\multicolumn{6}{@{}l}{\textit{(a) Arm Bistro, jittered, zero-shot}} \\
Method & PSNR-Y$\,\uparrow$ & SSIM-Y$\,\uparrow$ & LPIPS$\,\downarrow$ & FLIP$\,\downarrow$ & JOD$\,\uparrow$ \\
\midrule
Bicubic & 22.57 & 0.7743 & 0.269 & 0.1576 & 7.22 \\
Bicubic, registered & 24.21 & 0.8283 & 0.259 & 0.1241 & 7.70 \\
Lanczos-3 & 22.41 & 0.7662 & 0.291 & 0.1575 & 7.17 \\
TAAU (classical) & 27.16 & 0.9075 & 0.111 & 0.0922 & 8.58 \\
TAAU, clamp (classical) & 27.23 & 0.9103 & 0.107 & 0.0922 & 8.61 \\
Ours, single frame (45k) & 26.64 & 0.8726 & 0.126 & 0.1174 & 8.22 \\
Ours, recurrent (45k) & 31.07 & 0.9506 & 0.040 & 0.0764 & 9.23 \\
Ours, final (100k) & \textbf{31.62} & \textbf{0.9547} & \textbf{0.034} & \textbf{0.0752} & \textbf{9.28} \\
\midrule
\multicolumn{6}{@{}l}{\textit{(b) STSS (no jitter), PSNR-Y}} \\
Method & Arena$^{\ddagger}$ & Lewis & SunTemp & Subway & Mean \\
\midrule
STSSNet$^{\dagger}$ & 33.36 & 35.02 & \textbf{32.18} & \textbf{31.00} & 32.89 \\
ExtraNet+NSRR$^{\dagger}$ & 33.14 & 34.29 & 31.45 & 30.18 & 32.27 \\
RIFE+NSRR$^{\dagger}$ & 33.26 & 34.28 & 31.41 & 29.93 & 32.22 \\
Bicubic & 33.23 & 32.87 & 31.40 & 29.88 & 31.84 \\
TAAU (classical) & 31.35 & 32.83 & 30.89 & 30.12 & 31.30 \\
Ours, $\lambda_T=0$ & \textbf{33.90} & \textbf{37.49} & 31.58 & 30.96 & \textbf{33.48} \\
Ours, $\lambda_T=0.05$ & 33.11 & 37.40 & 31.55 & 30.94 & 33.25 \\
Ours, $\lambda_T=0.25$ & 30.26 & 37.09 & 29.62 & 29.24 & 31.55 \\
\bottomrule
\end{tabular}
\par\smallskip
\begin{minipage}{\linewidth}\fontsize{6.8}{8}\selectfont
$^{\dagger}$Published by He et al.~\cite{he2024low} (Table~1, super-resolved frames), one model per scene for Lewis, SunTemple and Subway and one trained on all three for Arena. Our STSS models are trained on the first Lewis training file only (2798 frames: about half of STSSNet's Lewis training set and 16\% of its three-scene set), so SunTemple, Subway and Arena are unseen scenes for us. $^{\ddagger}$Arena is unseen for every method. Scores use their metric code: even frame indices, 8-bit BT.601 luma, full frame; Section~\ref{sec:generality} quotes Arena on our protocol (all frames), hence slightly different values.
\end{minipage}
\end{table}
%% ======== END TABLE 3 (T7_bench) ========

\subsection{Ablations}\label{sec:ablations}

%% ======== TABLE 4 (main ablations) ========
\begin{table}[!t]
\centering
\caption{Accumulation, history, registration and temporal-weight ablations. Panel (c) removes individual registration corrections; the other panels use full registration. Panel (d) averages two seeds.}
\label{tab:ablation}
\fontsize{7.4}{8.6}\selectfont
\setlength{\tabcolsep}{2.0pt}
\begin{tabular*}{\columnwidth}{@{\extracolsep{\fill}}l c c c c@{}}
\toprule
Variant & Seed & PSNR-Y$\uparrow$ & LPIPS$\downarrow$ & STD$_{\mathrm s}$\\
\midrule
\multicolumn{5}{@{}l}{\textit{(a) Ours, $\times$4, 45k, $\lambda_T=0.05$}}\\
Normalised, 30 frames & 0 & 29.74 & 0.196 & 0.43\\
Single predicted blend & 0 & 29.71 & 0.195 & 0.53\\
Normalised, 16 frames & 0 & 29.60 & 0.208 & 0.33\\
\midrule
\multicolumn{5}{@{}l}{\textit{(b) History, 45k, $\lambda_T=0.05$, 30 frames}}\\
$\times$2, recurrent & --- & 32.58 & 0.095 & 0.11\\
$\times$2, no history & --- & 30.93 & 0.170 & 1.09\\
$\times$4, recurrent & --- & 29.74 & 0.196 & 0.44\\
$\times$4, no history & --- & 27.71 & 0.320 & 1.74\\
\midrule
\multicolumn{5}{@{}l}{\textit{(c) Registration, $\times$4, 15k, $\lambda_T=0$, 16 frames}}\\
Full registration & 0 & 29.31 & 0.203 & 0.83\\
Full registration & 1 & 29.38 & 0.223 & 0.74\\
Without sample phase & 0 & 29.22 & 0.208 & 0.81\\
Without registered skip & 0 & 29.28 & 0.205 & 0.80\\
Without LR re-jitter & 0 & 29.19 & 0.210 & 0.81\\
Without unjittered depth & 0 & div. & div. & div.\\
Without unjittered depth & 1 & 29.23 & 0.209 & 0.76\\
\midrule
\multicolumn{5}{@{}l}{\textit{(d) QRISP-L architecture, $\times$4, 45k, two-seed mean}}\\
$\lambda_T=0$ & 0,1 & 29.04 & 0.211 & 1.14\\
$\lambda_T=0.05$ & 0,1 & 28.98 & 0.233 & 0.61\\
\bottomrule
\end{tabular*}
\par\smallskip
\begin{minipage}{\linewidth}\fontsize{6.8}{8}\selectfont
Ours: 192 channels, eight blocks. STD: School frames 275--292, float output in (a,c), 8-bit in (b,d).
No history jointly removes colour history, accumulator and recurrent features; its seed is unreported in Supplementary Table~\ref{supp:tab:extra-rows}.
div.: non-finite from about 11k training iterations; seed 1 is compared with its own control.
Panel (d) retrains the QRISP-L architecture in our pipeline. See Supplementary Tables~\ref{supp:tab:ablation}, \ref{supp:tab:loo-full}, \ref{supp:tab:transfer} and \ref{supp:tab:extra-rows}.
\end{minipage}
\end{table}

%% ======== BEGIN TEXT 07e_ablations ========
Table~\ref{tab:ablation} presents four controlled comparison sets.
The 45k recipe comparison changes one accumulation or clip-length choice at a time.
Replacing normalised accumulation by a directly predicted blend gives comparable PSNR (29.71 versus
29.74\,dB) and nearly identical temporal PSNR, but increases float-output static STD from 0.43 to 0.53.
The accumulated weight reduces variation in this comparison; it contributes little PSNR.
Reducing the training clip from 30 to 16 frames lowers PSNR by 0.14\,dB and median static $\alpha$ from
$2.3\times10^{-3}$ to $8.3\times10^{-4}$.
The lower STD accompanies reduced sample uptake and PSNR; the longer clips improve both deployed sample uptake and reconstruction accuracy.

The no-history variants remove the accumulator, reprojected colour input and recurrent features together.
At 45k, their unrounded PSNR drops by 1.65\,dB at $\times$2 and 2.02\,dB at $\times$4, with higher LPIPS and
static STD (Table~\ref{tab:ablation}b). These runs measure the joint value of history; panel (a) isolates the accumulator.

Registration leave-one-out runs use $\lambda_T=0$, 15k iterations and 16-frame clips.
Removing phase, registered skip or LR re-jitter costs 0.09, 0.03 or 0.12\,dB against seed 0.
Without unjittered depth, seed 0 becomes non-finite at about 11k iterations; seed 1 loses 0.14\,dB against
its own control (panel c). The 0.06\,dB difference between control seeds before rounding is comparable to the phase and skip effects.
The motion-only variant loses 0.10\,dB and removing float32 accumulation loses 0.01\,dB
(Supplementary Table~\ref{supp:tab:loo-full}). The controls measure reconstruction, and the R/J grid measures the response to temporal weight across registration conventions.

Panel (d) applies the small weight to QRISP-L's re-implemented architecture with registered inputs,
averaged over two seeds at 45k. STD falls from 1.14 to 0.61 at 0.06\,dB lower PSNR and 0.17\,dB lower static PSNR, while LPIPS increases.

%% ======== END TEXT 07e_ablations ========

\subsection{Latency}\label{sec:latency}
% floats defined before the text so that they take the next column top

%% ======== BEGIN TEXT 07f_latency ========
Table~\ref{tab:headline}'s ms column measures one complete recurrent step at $\times$4, 1080p output and batch 1
on an H200, including input preparation, warps, network and accumulation, excluding rendering.
Every row uses fused kernels, bfloat16 with float32 accumulation, and one CUDA graph; we average 300 calls
after 50 warm-up calls. QRISP configurations use the same harness with random weights, paired with published
trained-model PSNR. Supplementary Table~\ref{supp:tab:latency} repeats these timings with parameter and GFLOP counts.

The $96\times4$ model reaches 29.49\,dB at 0.524\,ms, exceeding QRISP-L's published 29.42\,dB at 0.698\,ms
by 0.07\,dB with 25\% lower latency. The shift-MLP variants reach 29.32 and 29.35\,dB at 0.201 and
0.301\,ms: 0.10 and 0.07\,dB below QRISP-L at 71\% and 57\% lower latency. Both also exceed QRISP-M's
29.06\,dB at 0.443\,ms, by 0.26 to 0.29\,dB with 32\% to 55\% lower latency. 

The largest model reaches 30.04\,dB at 2.032\,ms, gaining 0.62\,dB over QRISP-L at 2.91 times its latency.
Within our variants, the six-block shift-MLP matches the $64\times4$ convolutional model's 29.32\,dB at
0.201 versus 0.332\,ms, a 39\% reduction. Increasing width from 192 to 256 adds 0.12\,dB for 0.774\,ms.

Registration corrects fields and offsets within existing warps, and the temporal weight affects training only. At a target of 120\,Hz on the H200, the $256\times8$ and $96\times4$ steps consume 24.4\% and 6.3\% of an 8.33\,ms frame budget. Rendering and other passes share the remainder, making the trunk size a practical quality--budget choice.
%% ======== END TEXT 07f_latency ========

%-------------------------------------------------------------------------

\section{Limitations}\label{sec:limitations}
%% ======== BEGIN TEXT 08_limitations ========
Stability is evaluated on three QRISP scenes; Arm adds one moving capture, and STSS uses separately retrained models with unjittered input. The unchanged QRISP model reaches 25.8\,dB on unjittered Arena, below bicubic's 33.2\,dB (Supplementary Section~\ref{supp:section-7}). Transfer therefore depends on the input jitter distribution. The mechanism grid is single-seed, with repeatability measured in separate 15k runs. Perceptual conclusions rely on image and video metrics.

Registration depends on motion describing the image change~\cite{zeng2021temporally}. At $\lambda_T=0$, R trails J by 0.13\,dB on SpaceShipDemo, largely in the particle sequence: sprites without motion vectors remain in the warped history. Reactive masks for particles and transparency~\cite{amd2022fsr2} are a possible extension; the QRISP buffers contain none. The oracle isolates motion; trained R/J runs compare all four registration conventions together.

The small weight trades some static accuracy and JOD for reduced variation on QRISP; the STSS runs achieve their highest PSNR at $\lambda_T=0$. Our strongest benchmark models also use more capacity than QRISP. Latency measurements cover the recurrent upscaling step on an H200. Profiling the complete renderer on consumer GPUs remains necessary for deployment.
%% ======== END TEXT 08_limitations ========

\section{Conclusion}\label{sec:conclusion}
%% ======== BEGIN TEXT 09_conclusion ========
Registration changes what a temporal derivative loss considers inexpensive. With accurate correspondence,
copying static history pays only the reference's small reprojected change and approximately retains the
existing reconstruction error. A large temporal weight can then suppress useful sample updates.
Jitter-carrying motion raises this copy cost and can conceal the behaviour.
Our oracle isolates the registration cost, and controlled training establishes the loss response across
three network families and two scales.
These findings motivate registering each state on its own lattice, checking correspondence without a network,
and reconsidering temporal weights after registration changes.
On QRISP, a small weight and normalised accumulation reduce static variation at a measured accuracy cost.
The resulting models improve published benchmark PSNR and transfer unchanged to Arm's jittered capture.
At $\times$4, our $96\times4$ variant gains 0.07\,dB over published QRISP-L PSNR at 25\% lower H200 latency.
Reporting sample uptake, accuracy and perceptual quality alongside stability metrics distinguishes useful
reconstruction from frozen history.
%% ======== END TEXT 09_conclusion ========

%-------------------------------------------------------------------------
\clearpage % references begin after the ten-page main paper
\bibliographystyle{eg-alpha-doi}
\bibliography{refs}

% Supplement figures, tables, equations and sections have independent S numbers.
\clearpage
\setcounter{section}{0}
\setcounter{subsection}{0}
\setcounter{figure}{0}
\setcounter{table}{0}
\setcounter{equation}{0}
\renewcommand{\thesection}{S\arabic{section}}
\renewcommand{\thefigure}{S\arabic{figure}}
\renewcommand{\thetable}{S\arabic{table}}
\renewcommand{\theequation}{S\arabic{equation}}
\renewcommand{\theHsection}{supp.\arabic{section}}
\renewcommand{\theHfigure}{supp.\arabic{figure}}
\renewcommand{\theHtable}{supp.\arabic{table}}
\renewcommand{\theHequation}{supp.\arabic{equation}}
\makeatletter
\gdef\@shorttitle{Frozen History -- Supplementary Material}
\makeatother
\twocolumn[{\centering\Large\bfseries Supplementary Material\par\vspace*{30pt}}]

% Supplement-specific spacing. The publication class and assets are unchanged.
\renewcommand{\topfraction}{0.95}
\renewcommand{\dbltopfraction}{0.95}
\renewcommand{\bottomfraction}{0.9}
\renewcommand{\textfraction}{0.05}
\renewcommand{\floatpagefraction}{0.95}
\renewcommand{\dblfloatpagefraction}{0.95}
\setcounter{topnumber}{4}
\setcounter{dbltopnumber}{4}
\setcounter{bottomnumber}{3}
\setcounter{totalnumber}{8}
\setlength{\textfloatsep}{8pt plus 1pt minus 1pt}
\setlength{\dbltextfloatsep}{8pt plus 1pt minus 1pt}
\setlength{\floatsep}{6pt plus 1pt minus 1pt}
\setlength{\dblfloatsep}{6pt plus 1pt minus 1pt}
\setlength{\intextsep}{7pt plus 1pt minus 1pt}
\setlength{\parskip}{3pt plus .5pt minus .5pt}
\setlength{\emergencystretch}{1em}
\makeatletter
\setlength{\@fptop}{0pt}
\setlength{\@fpsep}{8pt plus 1pt}
\setlength{\@fpbot}{0pt plus 1fil}
\setlength{\@dblfptop}{0pt}
\setlength{\@dblfpsep}{8pt plus 1pt}
\setlength{\@dblfpbot}{0pt plus 1fil}
% Match the class's caption style with shorter, fixed gaps.
\long\def\@makecaption#1#2{\vskip 3pt
 \setbox\@tempboxa\hbox{\textbf{#1:} \reset@font\textit{#2}}
 \ifdim \wd\@tempboxa>\hsize
   \textbf{#1:} \reset@font\itshape #2\par
 \else
   \hbox to\hsize{\hfil\box\@tempboxa\hfil}
 \fi
 \vskip 4pt}
\expandafter\let\expandafter\galfigure\csname figure*\endcsname
\expandafter\let\expandafter\endgalfigure\csname endfigure*\endcsname
\let\supp@table\table \let\supp@endtable\endtable
\renewenvironment{table}[1][]{\supp@table[!htbp]}{\supp@endtable}
\expandafter\let\expandafter\supp@tablestar\csname table*\endcsname
\expandafter\let\expandafter\supp@endtablestar\csname endtable*\endcsname
\renewenvironment{table*}[1][]{\supp@tablestar[!tbp]}{\supp@endtablestar}
\makeatother
\raggedbottom

\raggedbottom

Low temporal variation can accompany either sample accumulation or a copied history. We assess it together
with reconstruction accuracy, perceptual error and the deployed share of new samples.
All table differences are computed from unrounded scores and may differ from differences between the
displayed rounded values.

\section{Model-free copy cost on both lattices}\label{supp:section-1}
\begin{figure}[!htbp]
  \centering
  \includegraphics[width=\columnwidth]{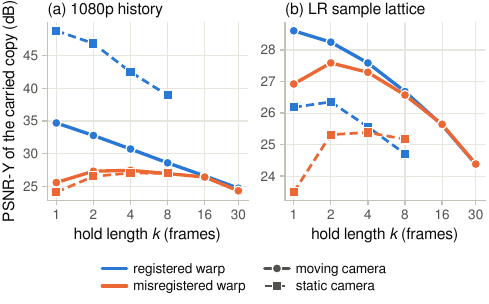}
  \caption{Model-free copy cost. PSNR after carrying a frame forward by $k$ chained reprojections: the high-resolution reference against the target reference on the 1080p output lattice (a), and the low-resolution render against the target render on the sample lattice (b), with registered (blue) and misregistered (orange) motion (jitter-carrying on the output lattice, unjittered on the sample lattice), for moving (solid) and static (dashed) camera frames. All hold lengths use a common set of valid pixels. No network is involved. On the output lattice, a one-frame copy reaches 48.7\,dB on the static-camera frames with registered motion, compared with 24.2\,dB with jitter-carrying motion. On the sample lattice the shipped jitter-carrying motion is the correct one.}
  \label{supp:fig:copy}
\end{figure}

Figure~\ref{supp:fig:copy} extends panel (d) of Figure~\ref{fig:pipeline} of the paper to the low-resolution sample lattice.
We use the unjittered motion field on the output lattice and the shipped jitter-carrying field on the sample lattice
(Section~\ref{sec:registration} of the paper, errors 1 and 2).
On the sample lattice, the mean absolute residual of the low-resolution frame carried one frame forward (in-bounds
pixels, all test frames) is 21\% larger with the unjittered field $m_t$ than with $\tilde m_t$ of
Equation~\ref{eq:lrfield} of the paper (0.0433 against 0.0357).
We report the model-free measurements used in Sections~\ref{sec:registration} and~\ref{sec:oracle}
of the paper in Table~\ref{supp:tab:modelfree}.

These checks isolate the geometry from the learned reconstruction. The motion field must transport values on
the lattice where those values are stored: removing jitter is appropriate for the output history, while the
sample lattice still needs the difference between successive sample offsets. The low-resolution residual
therefore tests a different requirement from the high-resolution copy cost.

On static-camera frames, the registered reference can be carried forward with little error even though no new
samples are accumulated. This establishes the low cost of copying before introducing a network. It does not
measure whether a trained model improves its reconstruction over time; we examine that question through the
sample share and copy fraction below. Using the same valid pixels at all hold lengths also keeps changes in
the comparison region from determining the copy-cost curves.

\begin{table*}[t]
\centering
\caption{Model-free measurements quoted in Sections~\ref{sec:registration} and~\ref{sec:oracle} of the paper
(QRISP test set, $\times$4 unless noted; no network involved).}
\label{supp:tab:modelfree}
\footnotesize
\setlength{\tabcolsep}{4pt}
\begin{tabular*}{\textwidth}{@{\extracolsep{\fill}}l l r@{}}
\toprule
Requirement & Measurement & Value \\
\midrule
(1) history on the output lattice & copy cost ($k=1$, $\ell_1$) on static pixels, $m$ vs.\ $\tilde m$ & $3.2\times10^{-4}$ vs.\ $3.76\times10^{-2}$ (118$\times$) \\
 & same on the longest camera stop / on moving frames & 1304$\times$ / 2.2$\times$ \\
 & held-copy PSNR ($k=1$, all frames), $m$ vs.\ $\tilde m$, $\times$4 / $\times$3 / $\times$2 & 35.71/25.45, 35.89/26.51, 36.09/28.30\,dB \\
(2) state on the input lattice & previous LR frame carried with $\tilde m$ (Equation~\ref{eq:lrfield}) vs.\ with $m$ & 28.35 vs.\ 26.52\,dB \\
(3) unjittered depth & mean $|$reprojected depth change$|$, static segment, jittered vs.\ unjittered & about $10^3\times$ \\
(4) registered skip & gain of the registered bicubic skip, moving SeaPort frames (5 frames) & $+0.86$ to $+1.53$\,dB \\
(4) sample phase & phases with the wrong row when $+j^y_t$ is used / mean row error & 16 of 16 / 1.80 output pixels \\
 & plain splat on the static segment: correct / mirrored phase / bicubic & 34.19 / 29.78 / 28.38\,dB \\
\midrule
Deployment & fused fp32 vs.\ reference implementation / bf16 vs.\ fp32 & $\le$0.0003 / $<$0.01\,dB \\
\bottomrule
\end{tabular*}
\end{table*}

\section{The registration $\times$ temporal-weight grid}\label{supp:section-2}
We report the full registration-by-temporal-weight grid in Table~\ref{supp:tab:grid}, corresponding to Figure~\ref{fig:dial}
and Section~\ref{sec:dial} of the paper. We compare the registered pipeline (R) with the misregistered pipeline (J),
which uses jitter-carrying motion on both lattices, jittered depth, an unregistered skip and a mirrored sample phase.
Both pipelines use jittered inputs.

Under R, temporal error and reconstruction accuracy respond differently to the weight. At the small weight,
PSNR remains close to the zero-weight control, while static standard deviation and the median share of new
samples decrease. At larger weights, temporal PSNR continues to rise, but PSNR and static JOD fall and FLIP
error increases. We use the sample-uptake measurements to check whether the lower temporal error accompanies
continued accumulation.

Under J, the same weight range produces much smaller changes in PSNR and static standard deviation. Both
pipelines receive jittered samples; their registration conventions determine how the motion, depth and sample
offsets enter the reconstruction.

%% ======== BEGIN tables/T2_grid ========
% T2_grid.tex -- generated by paper/make_tables.py (2026-10-01 15:56 UTC); do not edit by hand
% except the caption, which is preserved when the table is regenerated.
% Every number is read from the file named in the '% src:' comment above its row.
\begin{table}[t]
\centering
\caption{Registration $\times$ temporal weight: our $192\times8$ network at $\times$4, trained for 45k iterations.
Only the registration pipeline and $\lambda_T$ differ. Static standard deviation and sample share use the
AbandonedSchool stop; static JOD averages the three static segments. Temporal PSNR uses a shared unjittered
motion field. Metric definitions and averaging are given below.}
\label{supp:tab:grid}
\footnotesize
\setlength{\tabcolsep}{3.6pt}
\begin{tabular}{@{}l r r r r r r@{}}
\toprule
$\lambda_T$ & PSNR-Y$\uparrow$ & tPSNR$_{60}\!\uparrow$ & Std$_{\mathrm{s}}$ & JOD$_{\mathrm{s}}\!\uparrow$ & $\tilde\alpha_{\mathrm{s}}$ & FLIP$\downarrow$ \\
\midrule
\multicolumn{7}{@{}l}{\textit{R: registered pipeline}} \\
0 & 29.75 & 30.59 & 0.77 & 7.64 & 0.034 & 0.0801 \\
0.05 & 29.74 & 30.95 & 0.43 & 7.55 & $2.3{\times}10^{-3}$ & 0.0829 \\
0.25 & 29.48 & 31.21 & 0.17 & 7.26 & $3.2{\times}10^{-5}$ & 0.0881 \\
1 & 28.96 & 31.32 & 0.086 & 6.70 & $9.6{\times}10^{-7}$ & 0.0980 \\
\midrule
\multicolumn{7}{@{}l}{\textit{J: misregistered pipeline}} \\
0 & 29.72 & 30.15 & 1.36 & 7.58 & 0.61 & 0.0828 \\
0.05 & 29.78 & 30.23 & 1.35 & 7.60 & 0.50 & 0.0814 \\
0.25 & 29.81 & 30.28 & 1.33 & 7.61 & 0.46 & 0.0821 \\
1 & 29.74 & 30.16 & 1.40 & 7.56 & 0.39 & 0.0843 \\
\bottomrule
\end{tabular}
\par\smallskip
\begin{minipage}{\linewidth}\scriptsize
$\lambda_T$: temporal-loss weight. tPSNR$_{60}$: temporal PSNR of consecutive-frame differences warped with the non-jittered motion field, per-pair dB capped at 60, averaged. Std$_{\mathrm{s}}$: mean per-pixel temporal standard deviation, computed from floating-point output values and expressed in 8-bit intensity units, and $\tilde\alpha_{\mathrm{s}}$: median over pixels and frames of the deployed per-pixel new-sample share $\alpha$ (0 = history copied unchanged), both on AbandonedSchool/0001 frames 275--292 (18 frames). JOD$_{\mathrm{s}}$: ColorVideoVDP~\cite{mantiuk2024colorvideovdp} JOD averaged over the three stationary-camera segments of the test set. FLIP~\cite{andersson2020flip}: whole test set. J: trained and warped with the shipped jitter-carrying motion vectors on both lattices, with jittered depth, the unregistered skip and the mirrored sample phase (all registration flags off).
\end{minipage}
\end{table}
%% ======== END tables/T2_grid ========

\section{The temporal weight on a static view}\label{supp:section-3}
\begin{figure*}[!tbp]
  \centering
  \includegraphics[width=\textwidth]{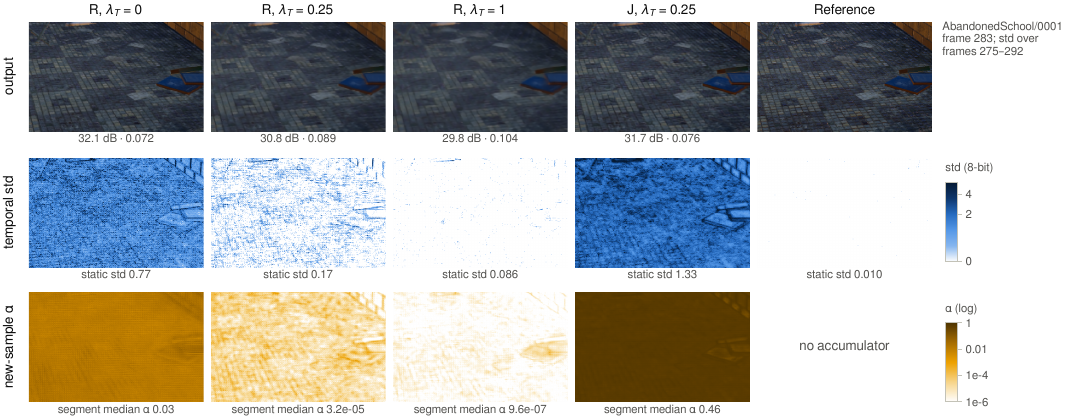}
  \caption{The static AbandonedSchool stop ($\times$4) for the R models with
    $\lambda_T = 0$, $0.25$ and $1$ and the J model with $\lambda_T = 0.25$: output crops at frame 283,
    temporal standard-deviation maps over frames 275--292 (8-bit intensity units), and per-pixel median deployed new-sample
    shares $\alpha$ over the same frames. Numbers below the output crops are crop PSNR-Y and \reflectbox{F}LIP; static STD
    (float output) and segment median $\alpha$ are full-frame segment statistics as in Table~\ref{supp:tab:grid}.}
  \label{supp:fig:supp-freeze}
\end{figure*}

Figure~\ref{supp:fig:supp-freeze} compares the R models at $\lambda_T=0$, $0.25$ and $1$ with the J model at
$\lambda_T=0.25$ on the AbandonedSchool stop. As $\lambda_T$ increases under R, temporal standard deviation decreases and
the deployed share of new samples decreases by several orders of magnitude, while the crops lose detail. The J model
retains higher temporal variation.

We compare the locations with low temporal standard deviation against the new-sample maps and reference crop.
Under R, larger weights reduce temporal standard deviation and sample uptake, while crop PSNR-Y falls
and FLIP error rises. The reference comparison shows the spatial cost of this increasingly constant output.

\section{Training dynamics of the freeze}\label{supp:section-4}
\begin{figure}[!htbp]
  \centering
  \includegraphics[width=\columnwidth]{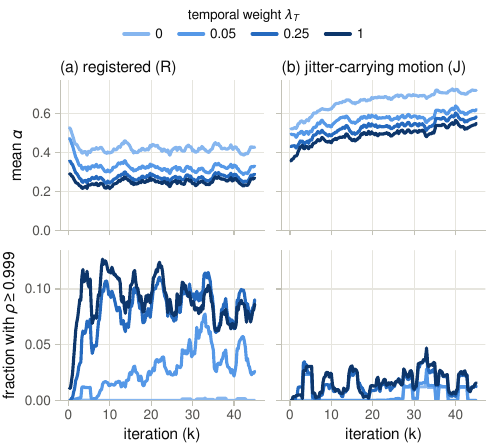}
  \caption{Training dynamics on the grid of Table~\ref{supp:tab:grid}. Top: mean new-sample share $\alpha$ on training crops.
Bottom: fraction with $\rho\geq0.999$. Left: R; right: J. These training averages are distinct from the
deployed static-segment medians.}
  \label{supp:fig:supp-alpha}
\end{figure}

Figure~\ref{supp:fig:supp-alpha} shows, for the registration $\times$ temporal-weight grid (Table~\ref{supp:tab:grid}), the
average share $\alpha$ of new samples and the fraction of pixels whose retention saturates on the training crops.
Over the last 10\% of training, the fraction of pixels with $\rho\geq0.999$ averages 4 to 9\% under R for
$\lambda_T\geq0.05$, compared with at most 2\% under J. These training-crop averages differ from the deployed
static-segment medians reported in Table~\ref{supp:tab:grid}.

We also train the R model for 15k iterations at $\lambda_T\in\{0,0.05,0.1,0.25\}$. In that order, PSNR is 29.31,
29.28, 29.14 and 28.82\,dB; static pixel standard deviation is 0.83, 0.41, 0.23 and 0.077; and deployed median $\alpha$
is 0.037, 0.0011, $<10^{-4}$ and $<10^{-4}$. The fractions of training pixels with saturated retention, averaged over
the last 10\% of training, are 0, 0.063, 0.103 and 0.146, respectively.

The training curves and deployed static statistics describe different populations. The curves average over
training crops, whereas the deployed median is measured during a fixed camera stop. A nonzero training mean
can coexist with a very small deployed static median; the two values are not alternative estimates of the same
quantity. The saturation fraction measures how often retention approaches its upper limit; the sample-share
statistic measures the resulting contribution of the newest sample.

The shorter 15k comparison retains the ordering seen in the grid: stronger temporal regularisation lowers
static variation and sample uptake while reducing PSNR. Its saturation fractions belong to that training
length and are reported separately from those of the 45k grid.

\section{Architecture, scale and loss comparisons}\label{supp:section-5}
We summarise the comparisons of Section~\ref{sec:generality} of the paper in Table~\ref{supp:tab:generality}
and report each run in Table~\ref{supp:tab:gen-full}. These are the measurements used in panels (c) and (d)
of Figure~\ref{fig:dial} of the paper. The settings differ
from the grid only where stated.
\begin{itemize}
\item \emph{Unwarped temporal term}: $\Delta x_t = x_t - x_{t-1}$ for output and reference, with every pixel valid;
  the objective and its normalisation are those of the warped term. In a zero-motion implementation check on random
  inputs with the same valid pixels and normalisation, the total loss values differ by $6\times10^{-8}$ ($\ell_1$) and
  $7\times10^{-8}$ (squared).
\item \emph{Direct prediction}: the network has no accumulator, no blend weight and no retention. We feed all 48
  channels of the warped previous output to the first convolution at $\times$4, without the accumulating network's
  16-channel projection, so that the network can represent a copy of the history; the output is the colour
  head plus the registered bicubic skip and becomes the next frame's history. At initialisation it equals the
  registered bicubic upsampling.
\item \emph{Copy fraction}: over AbandonedSchool/0001 frames 275--292, the fraction of in-bounds static pixels whose
  output differs from its own previous output, reprojected with the shared unjittered field, by less than $0.5/255$ in
  every RGB channel; static pixels are those where the reference itself changes by less than this amount, so the
  reference scores 1.
\end{itemize}
The temporal stability of the direct-prediction network (R, $\lambda_T=0.25$, $\times$4) depends mainly on its colour
history. In a one-step knockout at each frame of the static segment, replacing that history by the frame's registered
bicubic upsampling raises static pixel standard deviation from 0.55 to 2.79; zeroing only the recurrent features raises
it to 0.70. We check bitwise agreement with the grid-training code for one default configuration (R,
$\lambda_T=0.25$, $\times$4): initial weights, loss values and gradients, forward and backward passes, and the first two
training iterations (batch 2, 6-frame clips), including the optimiser state.

Table~\ref{supp:tab:split} splits the PSNR change of every setting into the static-camera frames and the moving frames of the
test set.

For the warped $\ell_1$ term, increasing the weight lowers PSNR and static JOD under R and increases copying
in both blending architectures, in direct prediction and at $\times$2. Under J, PSNR increases slightly and
the copy fraction remains lower. Direct prediction has no explicit accumulator, so its increase in copying
shows that the behaviour does not require an accumulator or a learned blend weight. The colour-history
knockout further identifies the input on which its static stability depends.

The alternative loss forms do not produce identical outcomes. At $\lambda_T=0.25$, the unwarped term raises
overall R PSNR by 0.09\,dB and increases copying, while lowering static-frame PSNR by 0.21\,dB. The squared term changes the R copy
fraction much less, from 0.85 to 0.87. Scaling it to the median logged $\ell_1$ loss value controls the loss-value
magnitude, without assuming that the two forms have identical gradients. The STSS row reports an accuracy
cost on unjittered input; it does not independently establish the same freezing mechanism.

The static and moving split explains why an aggregate score can conceal the static cost. The moving subset
contains many more frames, and the overall score averages by scene rather than by the number of frames in each
subset. The overall column in Table~\ref{supp:tab:split} therefore uses different weights from a
frame-count-weighted average of the two displayed subset columns.

%% ======== BEGIN tables/T3_generality ========
% T3_generality.tex -- generated by paper/make_tables.py (2026-10-01 15:56 UTC); do not edit by hand
% except the caption, which is preserved when the table is regenerated.
% Every number is read from the file named in the '% src:' comment above its row.
\begin{table}[t]
\centering
\caption{Architecture, scale and loss comparisons corresponding to Figure~\ref{fig:dial}c,d of the paper.
Changes are from $\lambda_T=0$ to $0.25$ under J and R. The first four rows use the warped $\ell_1$ term;
the remaining rows test other loss forms and unjittered STSS input.}
\label{supp:tab:generality}
\footnotesize
\setlength{\tabcolsep}{2.2pt}
\begin{tabular}{@{}l r r r r r r@{}}
\toprule
 & \multicolumn{2}{c}{$\Delta$PSNR-Y} & \multicolumn{2}{c}{$\Delta$JOD$_{\mathrm s}$} & \multicolumn{2}{c}{copy fraction} \\
\cmidrule(lr){2-3}\cmidrule(lr){4-5}\cmidrule(l){6-7}
Setting & J & R & J & R & J & R \\
\midrule
%      (difference of the two numbers above)
%      (difference of the two numbers above)
%      (difference of the two numbers above)
%      (difference of the two numbers above)
Ours & +0.09 & $-$0.27 & +0.03 & $-$0.38 & 0.20$\rightarrow$0.23 & 0.85$\rightarrow$0.99 \\
%      (difference of the two numbers above)
%      (difference of the two numbers above)
%      (difference of the two numbers above)
%      (difference of the two numbers above)
Ours, $\times$2 & +0.20 & $-$0.09 & +0.01 & $-$0.14 & 0.32$\rightarrow$0.45 & 0.87$\rightarrow$1.00 \\
%      (difference of the two numbers above)
%      (difference of the two numbers above)
%      (difference of the two numbers above)
%      (difference of the two numbers above)
QRISP-style$^{a}$ & +0.23 & $-$0.32 & +0.10 & $-$0.69 & 0.11$\rightarrow$0.16 & 0.70$\rightarrow$0.98 \\
%      (difference of the two numbers above)
%      (difference of the two numbers above)
%      (difference of the two numbers above)
%      (difference of the two numbers above)
Direct prediction & +0.11 & $-$0.31 & +0.01 & $-$0.34 & 0.21$\rightarrow$0.31 & 0.21$\rightarrow$0.77 \\
%      (difference of the two numbers above)
%      (difference of the two numbers above)
%      (difference of the two numbers above)
%      (difference of the two numbers above)
Unwarped term & +0.08 & +0.09 & $-$0.11 & $-$0.10 & 0.20$\rightarrow$0.55 & 0.85$\rightarrow$0.96 \\
%      (difference of the two numbers above)
%      (difference of the two numbers above)
%      (difference of the two numbers above)
%      (difference of the two numbers above)
Squared term$^{b}$ & +0.06 & $-$0.04 & 0.00 & $-$0.04 & 0.20$\rightarrow$0.19 & 0.85$\rightarrow$0.87 \\
%      (difference of the two numbers above)
%      (difference of the two numbers above)
STSS, no jitter$^{c}$ & --- & $-$3.70 & --- & $-$0.74 & --- & --- \\
\bottomrule
\end{tabular}
\par\smallskip
\begin{minipage}{\linewidth}\scriptsize
Changes from $\lambda_T=0$ to $\lambda_T=0.25$ under the J and R pipelines; $\times$4 and 45k iterations unless noted. JOD$_{\mathrm s}$: ColorVideoVDP on the three static segments. Copy fraction: share of in-bounds static pixels (AbandonedSchool/0001, frames 275--292) whose output differs from its reprojected previous output by less than $0.5/255$ in every RGB channel; the reference scores 1. $^{a}$15k iterations. $^{b}$At the same $\lambda_T$ the squared term is much weaker than the $\ell_1$ term. $^{c}$Arena, $\times$2, our protocol; whole-sequence JOD instead of JOD$_{\mathrm s}$.
\end{minipage}
\end{table}
%% ======== END tables/T3_generality ========

%% ======== BEGIN tables/T3s_generality_full ========
% T3s_generality_full.tex -- generated by paper/make_tables.py (2026-10-01 15:56 UTC); do not edit by hand
% except the caption, which is preserved when the table is regenerated.
% Every number is read from the file named in the '% src:' comment above its row.
\begin{table*}[t]
\centering
\caption{Every run behind Figure~\ref{fig:dial}c,d of the paper (R: registered pipeline, J: misregistered pipeline). For the warped $\ell_1$ term, raising $\lambda_T$ from 0 to 0.25 increases the copy fraction under R in all tested architectures and at both scales; the increase is smaller under J. The final block compares alternative loss forms.}
\label{supp:tab:gen-full}
\footnotesize
\setlength{\tabcolsep}{4pt}
\begin{tabular}{@{}l l r r r r r r r r r@{}}
\toprule
Run & & PSNR-Y & SSIM-Y & LPIPS & tPSNR$_{60}$ & Std$_{\mathrm s}$ & JOD & JOD$_{\mathrm s}$ & $\tilde\alpha_{\mathrm s}$ & copy \\
\midrule
\multicolumn{11}{@{}l}{\textit{Our network, $\times$4}} \\
R & $\lambda_T{=}0$ & 29.75 & 0.8560 & 0.177 & 30.59 & 0.769 & 7.71 & 7.64 & 0.034 & 0.855 \\
R & $\lambda_T{=}0.25$ & 29.48 & 0.8432 & 0.221 & 31.21 & 0.174 & 7.60 & 7.26 & $3.2\times10^{-5}$ & 0.986 \\
J & $\lambda_T{=}0$ & 29.72 & 0.8528 & 0.193 & 30.15 & 1.361 & 7.70 & 7.58 & 0.61 & 0.199 \\
J & $\lambda_T{=}0.25$ & 29.81 & 0.8571 & 0.179 & 30.28 & 1.332 & 7.71 & 7.61 & 0.46 & 0.234 \\
\multicolumn{11}{@{}l}{\textit{Our network, $\times$2}} \\
R & $\lambda_T{=}0$ & 32.56 & 0.9223 & 0.090 & 32.93 & 0.407 & 8.61 & 8.27 & 0.086 & 0.869 \\
R & $\lambda_T{=}0.25$ & 32.46 & 0.9192 & 0.102 & 33.65 & 0.025 & 8.57 & 8.14 & $2.1\times10^{-5}$ & 0.997 \\
J & $\lambda_T{=}0$ & 32.56 & 0.9221 & 0.091 & 32.58 & 0.874 & 8.60 & 8.28 & 0.64 & 0.324 \\
J & $\lambda_T{=}0.25$ & 32.76 & 0.9254 & 0.086 & 32.97 & 0.757 & 8.63 & 8.28 & 0.44 & 0.448 \\
\multicolumn{11}{@{}l}{\textit{QRISP-style network, $\times$4, 15k}} \\
R & $\lambda_T{=}0$ & 28.73 & 0.8215 & 0.241 & 29.61 & 1.250 & 7.34 & 7.28 & 0.031 & 0.705 \\
R & $\lambda_T{=}0.25$ & 28.41 & 0.7991 & 0.313 & 30.34 & 0.224 & 7.10 & 6.59 & $1.1\times10^{-5}$ & 0.979 \\
J & $\lambda_T{=}0$ & 28.32 & 0.7999 & 0.318 & 28.82 & 1.983 & 7.20 & 6.92 & 0.42 & 0.109 \\
J & $\lambda_T{=}0.25$ & 28.54 & 0.8115 & 0.273 & 29.03 & 1.864 & 7.25 & 7.02 & 0.31 & 0.159 \\
\multicolumn{11}{@{}l}{\textit{Direct prediction, $\times$4}} \\
R & $\lambda_T{=}0$ & 29.71 & 0.8543 & 0.183 & 30.34 & 1.289 & 7.69 & 7.59 & --- & 0.212 \\
R & $\lambda_T{=}0.25$ & 29.40 & 0.8386 & 0.232 & 31.23 & 0.552 & 7.56 & 7.25 & --- & 0.775 \\
J & $\lambda_T{=}0$ & 29.65 & 0.8505 & 0.190 & 30.22 & 1.363 & 7.67 & 7.59 & --- & 0.212 \\
J & $\lambda_T{=}0.25$ & 29.77 & 0.8552 & 0.179 & 30.39 & 1.238 & 7.68 & 7.59 & --- & 0.310 \\
\multicolumn{11}{@{}l}{\textit{Loss form, $\times$4 (controls: first block)}} \\
R & unwarped, $\lambda_T{=}0.25$ & 29.84 & 0.8595 & 0.173 & 30.71 & 0.375 & 7.72 & 7.54 & $2.9\times10^{-4}$ & 0.965 \\
J & unwarped, $\lambda_T{=}0.25$ & 29.80 & 0.8559 & 0.189 & 30.63 & 0.707 & 7.70 & 7.47 & 0.69 & 0.551 \\
R & squared, $\lambda_T{=}0.25$ & 29.71 & 0.8540 & 0.184 & 30.83 & 0.716 & 7.69 & 7.60 & 0.028 & 0.869 \\
J & squared, $\lambda_T{=}0.25$ & 29.78 & 0.8531 & 0.190 & 30.24 & 1.373 & 7.70 & 7.58 & 0.59 & 0.191 \\
R & unwarped, $\lambda_T{=}1$ & 29.81 & 0.8603 & 0.171 & 30.72 & 0.271 & 7.69 & 7.45 & $4.0\times10^{-5}$ & 0.978 \\
J & unwarped, $\lambda_T{=}1$ & 29.71 & 0.8553 & 0.190 & 30.55 & 0.718 & 7.67 & 7.35 & 0.71 & 0.606 \\
R & squared, $\lambda_T{=}2.35$ & 29.58 & 0.8462 & 0.211 & 31.20 & 0.564 & 7.61 & 7.47 & 0.031 & 0.884 \\
J & squared, $\lambda_T{=}2.35$ & 29.69 & 0.8528 & 0.184 & 30.12 & 1.447 & 7.67 & 7.59 & 0.49 & 0.187 \\
\bottomrule
\end{tabular}
\par\smallskip
\begin{minipage}{\linewidth}\scriptsize
45k iterations unless noted. Std$_{\mathrm s}$: per-pixel temporal standard deviation of the output, computed from floating-point values in 8-bit intensity units, on AbandonedSchool/0001 frames 275--292; $\tilde\alpha_{\mathrm s}$: deployed median share of the newest sample there (accumulating and blending networks only); copy: share of in-bounds static pixels whose output differs from its reprojected previous output by less than $0.5/255$ in every RGB channel. At $\lambda_T=2.35$ the squared term is scaled to the median logged loss value of the $\ell_1$ term at $\lambda_T=0.25$ over the second half of training (ratio 9.4, measured on the $\lambda_T=0.25$ runs).
\end{minipage}
\end{table*}
%% ======== END tables/T3s_generality_full ========

%% ======== BEGIN tables/T3s_split ========
% T3s_split.tex -- generated by paper/make_tables.py (2026-10-01 15:56 UTC); do not edit by hand
% except the caption, which is preserved when the table is regenerated.
% Every number is read from the file named in the '% src:' comment above its row.
\begin{table*}[t]
\centering
\caption{Change in PSNR-Y from zero temporal weight to the indicated weight under J and R, reported separately
for static-camera frames, moving frames and the complete test set.}
\label{supp:tab:split}
\footnotesize
\setlength{\tabcolsep}{4pt}
\begin{tabular*}{\textwidth}{@{\extracolsep{\fill}}l c r r r r r r@{}}
\toprule
 & & \multicolumn{3}{c}{J} & \multicolumn{3}{c}{R} \\
\cmidrule(lr){3-5}\cmidrule(l){6-8}
Setting & $\lambda_T$ & static (97) & moving (1363) & all & static (97) & moving (1363) & all \\
\midrule
%      (difference of the two numbers above)
%      (difference of the two numbers above)
%      (difference of the two numbers above)
%      (difference of the two numbers above)
%      (difference of the two numbers above)
%      (difference of the two numbers above)
Ours, $\times$4 & $0\rightarrow0.25$ & +0.11 & +0.10 & +0.09 & $-$0.86 & $-$0.24 & $-$0.27 \\
%      (difference of the two numbers above)
%      (difference of the two numbers above)
%      (difference of the two numbers above)
%      (difference of the two numbers above)
%      (difference of the two numbers above)
%      (difference of the two numbers above)
Ours, $\times$4 & $0\rightarrow0.05$ & +0.05 & +0.07 & +0.07 & $-$0.19 & 0.00 & $-$0.02 \\
%      (difference of the two numbers above)
%      (difference of the two numbers above)
%      (difference of the two numbers above)
%      (difference of the two numbers above)
%      (difference of the two numbers above)
%      (difference of the two numbers above)
Ours, $\times$4 & $0\rightarrow1$ & +0.02 & +0.04 & +0.02 & $-$1.98 & $-$0.74 & $-$0.80 \\
%      (difference of the two numbers above)
%      (difference of the two numbers above)
%      (difference of the two numbers above)
%      (difference of the two numbers above)
%      (difference of the two numbers above)
%      (difference of the two numbers above)
Ours, $\times$2 & $0\rightarrow0.25$ & +0.23 & +0.20 & +0.20 & $-$0.68 & $-$0.05 & $-$0.09 \\
%      (difference of the two numbers above)
%      (difference of the two numbers above)
%      (difference of the two numbers above)
Ours, $\times$2$^{a}$ & $0\rightarrow0.05$ & --- & --- & --- & $-$0.20 & +0.04 & +0.02 \\
%      (difference of the two numbers above)
%      (difference of the two numbers above)
%      (difference of the two numbers above)
%      (difference of the two numbers above)
%      (difference of the two numbers above)
%      (difference of the two numbers above)
QRISP-style$^{b}$ & $0\rightarrow0.25$ & +0.31 & +0.23 & +0.23 & $-$0.85 & $-$0.30 & $-$0.32 \\
%      (difference of the two numbers above)
%      (difference of the two numbers above)
%      (difference of the two numbers above)
%      (difference of the two numbers above)
%      (difference of the two numbers above)
%      (difference of the two numbers above)
Direct prediction & $0\rightarrow0.25$ & +0.12 & +0.12 & +0.11 & $-$0.70 & $-$0.29 & $-$0.31 \\
%      (difference of the two numbers above)
%      (difference of the two numbers above)
%      (difference of the two numbers above)
%      (difference of the two numbers above)
%      (difference of the two numbers above)
%      (difference of the two numbers above)
Unwarped term & $0\rightarrow0.25$ & $-$0.15 & +0.13 & +0.08 & $-$0.21 & +0.14 & +0.09 \\
%      (difference of the two numbers above)
%      (difference of the two numbers above)
%      (difference of the two numbers above)
%      (difference of the two numbers above)
%      (difference of the two numbers above)
%      (difference of the two numbers above)
Unwarped term & $0\rightarrow1$ & $-$0.53 & +0.06 & 0.00 & $-$0.53 & +0.13 & +0.06 \\
%      (difference of the two numbers above)
%      (difference of the two numbers above)
%      (difference of the two numbers above)
%      (difference of the two numbers above)
%      (difference of the two numbers above)
%      (difference of the two numbers above)
Squared term & $0\rightarrow0.25$ & +0.03 & +0.07 & +0.06 & $-$0.09 & $-$0.04 & $-$0.04 \\
%      (difference of the two numbers above)
%      (difference of the two numbers above)
%      (difference of the two numbers above)
%      (difference of the two numbers above)
%      (difference of the two numbers above)
%      (difference of the two numbers above)
Squared term$^{c}$ & $0\rightarrow2.35$ & 0.00 & $-$0.04 & $-$0.03 & $-$0.37 & $-$0.16 & $-$0.18 \\
\bottomrule
\end{tabular*}
\par\smallskip
\begin{minipage}{\linewidth}\scriptsize
$\Delta$PSNR-Y from the $\lambda_T=0$ control of the same network and registration; $\times$4 and 45k iterations unless noted. Static and moving: plain means over the 97 static-camera frames (median motion of the unjittered 270p field below 0.02 pixels) and the 1363 moving frames of the test set; all: the test-set PSNR-Y (mean over the three scenes), as in Table~\ref{supp:tab:gen-full}. $^{a}$The 45k-iteration $\times$2 run with the final recipe against the $\times$2 grid control; the two differ only in $\lambda_T$ (registered only). $^{b}$15k iterations. $^{c}$Scaled to the median loss value of the $\ell_1$ term at $\lambda_T=0.25$ (Table~\ref{supp:tab:gen-full}).
\end{minipage}
\end{table*}
%% ======== END tables/T3s_split ========

\section{QRISP-style architecture and the STSS dataset}\label{supp:section-6}
We evaluate our re-implementation of the QRISP architecture with our data pipeline, losses and schedule for 15k
iterations. This tests the mechanism rather than reproducing published QRISP scores. Table~\ref{supp:tab:second} reports
these runs and the STSS experiment summarised in Section~\ref{sec:generality} of the paper.

The QRISP-style runs hold the architecture fixed while changing the pipeline and temporal weight. They use
a single-$\alpha$ blend at $\times$4 and the same 15k schedule, so their comparison tests the registration--loss
interaction without substituting our wider trunk. Raising the weight increases PSNR under J and decreases it
under R. Their training protocol differs from the published QRISP protocol, so we use them for this
interaction test rather than for a published-score comparison.

The STSS Arena runs ask a separate question on input rendered without jitter. The weight reduces PSNR by
3.70\,dB and lowers both whole-sequence and static temporal PSNR. This accuracy cost should be reported
separately from the copy-fraction evidence on QRISP, where the same static content can be transported almost
unchanged by registered motion.

%% ======== BEGIN tables/T3_second ========
% T3_second.tex -- generated by paper/make_tables.py (2026-10-01 15:56 UTC); do not edit by hand
% except the caption, which is preserved when the table is regenerated.
% Every number is read from the file named in the '% src:' comment above its row.
\begin{table}[t]
\centering
\caption{QRISP-style architecture and STSS Arena. (a) Single-$\alpha$ blend, $\times$4, 15k iterations, with our
pipeline and schedule. (b) Arena at $\times$2 without jitter. These are mechanism comparisons;
the QRISP-style scores do not reproduce the published QRISP protocol.}
\label{supp:tab:second}
\footnotesize
\setlength{\tabcolsep}{3.0pt}
\begin{tabular}{@{}l r r r r r r@{}}
\toprule
\multicolumn{7}{@{}l}{\textit{(a) QRISP-style architecture, QRISP, $\times$4, 15k}} \\
\midrule
Pipeline & $\lambda_T$ & PSNR-Y$\uparrow$ & tPSNR$_{60}\!\uparrow$ & Std$_{\mathrm{s}}$ & JOD$_{\mathrm{s}}\!\uparrow$ & $\tilde\alpha_{\mathrm{s}}$ \\
\midrule
%      (mean over static segments)
R & 0 & 28.73 & 29.61 & 1.25 & 7.28 & 0.031 \\
%      (mean over static segments)
R & 0.25 & 28.41 & 30.34 & 0.22 & 6.59 & $1.1{\times}10^{-5}$ \\
%      (mean over static segments)
J & 0 & 28.32 & 28.82 & 1.98 & 6.92 & 0.42 \\
%      (mean over static segments)
J & 0.25 & 28.54 & 29.03 & 1.86 & 7.02 & 0.31 \\
\midrule
\multicolumn{7}{@{}l}{\textit{(b) STSS Arena, $\times$2, no jitter, 45k}} \\
\midrule
Method & $\lambda_T$ & PSNR-Y$\uparrow$ & SSIM-Y$\uparrow$ & LPIPS$\downarrow$ & tPSNR$_{60}\!\uparrow$ & tPSNR$_{\mathrm{s}}\!\uparrow$ \\
\midrule
Bicubic & --- & 33.24 & 0.9433 & 0.116 & 34.05 & 40.40 \\
Ours & 0 & 34.16 & 0.9453 & 0.087 & 40.43 & 46.25 \\
Ours & 0.25 & 30.46 & 0.9317 & 0.124 & 39.09 & 36.70 \\
\bottomrule
\end{tabular}
\par\smallskip
\begin{minipage}{\linewidth}\scriptsize
(a) Same protocol, columns and R/J definitions as Table~\ref{supp:tab:grid}. (b) STSS~\cite{he2024low} Arena test scene (rendered without anti-aliasing), every frame of the 1001-frame test range after 8 warm-up frames; tPSNR$_{\mathrm{s}}$ pools the frame pairs with a stationary camera. 
\end{minipage}
\end{table}
%% ======== END tables/T3_second ========

\section{Other benchmarks: details}\label{supp:section-7}
\paragraph*{Arm Bistro capture.} The capture released with Arm's neural supersampling~\cite{arm2025nss} contains one Unreal Engine
camera run of 100 frames at 30\,Hz, of which 92 are scored after 8 warm-up frames, 540p to 1080p, with a period-32 jitter on a
1/8-pixel grid, jitter-free motion vectors, jittered depth and a 16-spp reference.

Because the supplied training and validation crops come from the same frames, we use this capture only for zero-shot
evaluation. All learned models are trained on QRISP, and the classical baselines retain their QRISP parameters. We
convert jitter, motion and depth to the registered conventions of Section~\ref{sec:registration} of the paper. The
model-free checks in Table~\ref{supp:tab:arm-checks} test the jitter sign, motion convention and two-lattice correction.

Input and reference are tone-mapped with the operator of Arm's own evaluation code. Table~\ref{supp:tab:arm-more} gives further rows.

\begin{figure*}[!tbp]
  \centering
  \includegraphics[width=\textwidth]{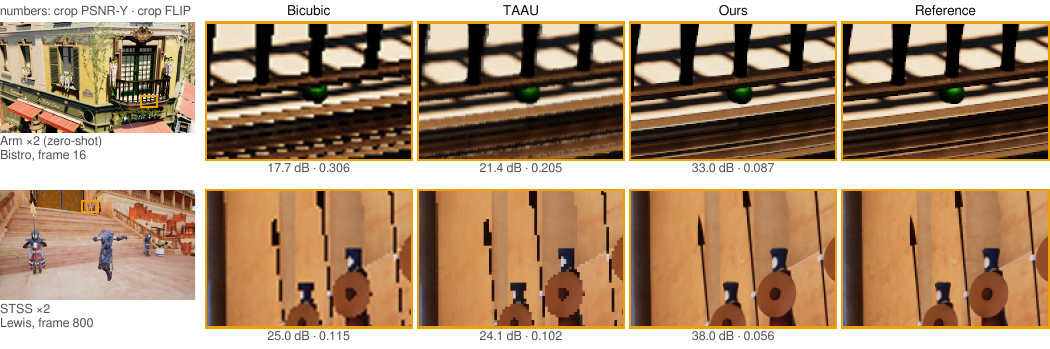}
  \caption{Selected $\times$2 crops on the Arm Bistro capture, zero-shot (top), and STSS Lewis with our STSS-trained
model at $\lambda_T=0.05$ (bottom): bicubic, TAAU, our model and the reference. Full reference frames are shown at the left; labels give crop PSNR-Y and \reflectbox{F}LIP.}
  \label{supp:fig:bench}
\end{figure*}

When each receives the conventions
it was trained with, the J $\times$2 model is within 0.07\,dB of the R model in PSNR at
$\lambda_T=0$ (31.06 against 31.12\,dB), and 0.36\,dB higher at $\lambda_T=0.25$ (31.32 against 30.96\,dB); at both
weights the R models lead in temporal PSNR, \reflectbox{F}LIP and JOD, and the J models in SSIM and
LPIPS. Feeding the J models the jitter-free motion field of the R models, or feeding our model a
mirrored jitter, no jitter or negated motion, costs 6 to 12\,dB against the same checkpoint with its own conventions. For the three methods evaluated in sRGB space (ours, TAAU and bicubic), the ordering is
unchanged in the reported metrics. Figure~\ref{supp:fig:bench} shows crops from the Arm capture and from STSS.

The Arm checks are performed before evaluating a learned model. They compare conventions using sample
alignment or frame transport in linear colour, so these errors are independent of network
reconstruction. The learned comparisons then use each model's own training conventions. Deliberately
changing those inputs tests convention sensitivity of the same checkpoint; it is a different comparison from
training R and J separately.

\paragraph*{STSS.} STSS has four test scenes (Arena, Lewis, SunTemple, Subway), rendered without
anti-aliasing; we verified that its samples carry no sub-pixel offset. Table~\ref{tab:bench}b of
the paper uses the metric code of He et al.~\cite{he2024low}; Table~\ref{supp:tab:stss-ours} repeats the comparison with our protocol and
graphics metrics. Our STSS models are trained on 16-frame clips, unlike the 30-frame clips of our QRISP recipe.
STSS does not provide jitter-induced sample diversity. On our protocol, TAAU falls below bicubic on two of the four
scenes, and neighbourhood clamping falls below it on three (Table~\ref{supp:tab:stss-ours}). Our QRISP-trained model performs
poorly when applied unchanged to the unjittered Arena sequence: PSNR falls from 34.6 to below 20\,dB during the sequence,
averaging 25.8\,dB against 33.2\,dB for bicubic upsampling, without producing non-finite values.

The external results also use different training conditions. Arm is a zero-shot capture for every learned
model. The STSS models are trained on the first Lewis training file with the shorter clips stated in the
table. The Lewis crop therefore illustrates an STSS-trained model, while the Bistro crop illustrates transfer
from QRISP. In the Bistro crop, bicubic and TAAU lose fine structure that the QRISP-trained model reconstructs.

%% ======== BEGIN tables/T7s_arm_checks ========
% T7s_arm_checks.tex -- generated by paper/make_tables.py (2026-10-01 15:56 UTC); do not edit by hand
% except the caption, which is preserved when the table is regenerated.
% Every number is read from the file named in the '% src:' comment above its row.
\begin{table}[t]
\centering
\caption{Model-free checks of jitter sign, motion convention and the two-lattice correction on the Arm capture.
The convention used gives the lowest error among the tested alternatives in each check.}
\label{supp:tab:arm-checks}
\footnotesize
\begin{tabular}{@{}l r@{}}
\toprule
\multicolumn{2}{@{}l}{\textit{V1. Jitter: $|$LR $-$ reference at the sample positions$|$}} \\
convention used & 0.0248 \\
$y$ mirrored & 0.0455 \\
$x$ mirrored & 0.0415 \\
axes swapped & 0.0433 \\
zero jitter & 0.0385 \\
\multicolumn{2}{@{}l}{\textit{V2. Motion: $|$warped previous reference $-$ reference$|$}} \\
convention used & 0.0105 \\
no motion & 0.0544 \\
negated & 0.0794 \\
$y$ negated & 0.0204 \\
axes swapped & 0.0758 \\
\multicolumn{2}{@{}l}{\textit{V3. Two lattices: $|$warped previous LR frame $-$ LR frame$|$}} \\
field $+\,\Delta o$ (used) & 0.0418 \\
field only & 0.0562 \\
field $-\,\Delta o$ & 0.0761 \\
no motion & 0.0771 \\
\bottomrule
\end{tabular}
\par\smallskip
\begin{minipage}{\linewidth}\scriptsize
V1: jitter check; V2: motion check; V3: two-lattice check. Mean absolute error in linear colour over all frames (V1) or over consecutive frame pairs (V2, V3); lower is better. The conventions we use give the lowest error in every check.
\end{minipage}
\end{table}
%% ======== END tables/T7s_arm_checks ========

%% ======== BEGIN tables/T7s_arm_more ========
% T7s_arm_more.tex -- generated by paper/make_tables.py (2026-10-01 15:56 UTC); do not edit by hand
% except the caption, which is preserved when the table is regenerated.
% Every number is read from the file named in the '% src:' comment above its row.
\begin{table}[t]
\centering
\caption{Additional Arm Bistro results at $\times$2, zero-shot. R and J use their own conventions unless noted.
Wrong-convention runs lose 6.1 to 12.0\,dB relative to the same checkpoint with its own conventions.
The checkpoint and input changes are specified below.}
\label{supp:tab:arm-more}
\footnotesize
\setlength{\tabcolsep}{1.6pt}
\begin{tabular}{@{}l r r r r r r@{}}
\toprule
Run & PSNR & SSIM & LPIPS & FLIP & JOD & tPSNR \\
\midrule
R, $\lambda_T{=}0$ ($\times$2 grid) & 31.12 & 0.9476 & 0.041 & 0.0755 & 9.22 & 32.94 \\
R, $\lambda_T{=}0.25$ & 30.96 & 0.9490 & 0.044 & 0.0803 & 9.20 & 33.78 \\
J, $\lambda_T{=}0$, own conv. & 31.06 & 0.9491 & 0.040 & 0.0845 & 9.12 & 32.10 \\
J, $\lambda_T{=}0.25$, own conv. & 31.32 & 0.9517 & 0.038 & 0.0814 & 8.99 & 32.84 \\
J, $\lambda_T{=}0$, jitter-free motion & 23.93 & 0.8081 & 0.170 & 0.1503 & 7.29 & 24.28 \\
% h3_x2_100k [encode=srgb]: flip not computed (flag in the run's argv; no non-finite frame) -> '---'
Ours, sRGB space & 30.53 & 0.9565 & 0.042 & --- & 9.10 & 32.35 \\
% taau [encode=srgb]: flip not computed (flag in the run's argv; no non-finite frame) -> '---'
TAAU, sRGB space & 26.60 & 0.9128 & 0.116 & --- & 8.43 & 29.05 \\
% bicubic [encode=srgb]: flip not computed (flag in the run's argv; no non-finite frame) -> '---'
Bicubic, sRGB space & 22.49 & 0.7927 & 0.285 & --- & 7.25 & 19.11 \\
% h3_x2_100k [jitter=ymirror]: flip not computed (flag in the run's argv; no non-finite frame) -> '---'
% h3_x2_100k [jitter=ymirror]: jod not computed (flag in the run's argv; no non-finite frame) -> '---'
Ours, jitter $y$ mirrored & 23.86 & 0.7945 & 0.142 & --- & --- & 27.26 \\
% h3_x2_100k [jitter=zero]: flip not computed (flag in the run's argv; no non-finite frame) -> '---'
% h3_x2_100k [jitter=zero]: jod not computed (flag in the run's argv; no non-finite frame) -> '---'
Ours, zero jitter & 19.59 & 0.6539 & 0.289 & --- & --- & 27.24 \\
% h3_x2_100k [motion=negate]: flip not computed (flag in the run's argv; no non-finite frame) -> '---'
% h3_x2_100k [motion=negate]: jod not computed (flag in the run's argv; no non-finite frame) -> '---'
Ours, motion negated & 25.50 & 0.8483 & 0.125 & --- & --- & 25.65 \\
\bottomrule
\end{tabular}
\par\smallskip
\begin{minipage}{\linewidth}\scriptsize
J models are trained with QRISP's shipped conventions; `own conv.' feeds them the jitter-carrying motion $m+(o_t-o_{t-1})$ and jittered depth; `jitter-free motion' feeds them the jitter-free motion field that the R models receive, with depth and jitter unchanged. `Ours': the $\times$2 model of Table~\ref{tab:bench} of the paper (192$\times$8, 100k iterations, $\lambda_T=0.05$, 31.62\,dB with its own conventions); the sRGB and wrong-convention rows use this checkpoint. PSNR and SSIM on luma; tPSNR capped at 60\,dB per pair. The sRGB rows re-encode input and reference after the tone map; the last three rows feed wrong conventions on purpose. ---: not computed.
\end{minipage}
\end{table}
%% ======== END tables/T7s_arm_more ========

%% ======== BEGIN tables/T7s_stss_ours ========
% T7s_stss_ours.tex -- generated by paper/make_tables.py (2026-10-01 15:56 UTC); do not edit by hand
% except the caption, which is preserved when the table is regenerated.
% Every number is read from the file named in the '% src:' comment above its row.
\begin{table*}[t]
\centering
\caption{STSS ($\times$2, no jitter) on our protocol, per test scene. Compared with $\lambda_T=0$, $\lambda_T=0.25$ lowers PSNR on all four scenes, with the largest drop of 3.70\,dB on Arena.}
\label{supp:tab:stss-ours}
\footnotesize
\setlength{\tabcolsep}{3.2pt}
\begin{tabular}{@{}l r r r r r r r r r r r r @{}}
\toprule
 & \multicolumn{3}{c}{Arena} & \multicolumn{3}{c}{Lewis} & \multicolumn{3}{c}{SunTemple} & \multicolumn{3}{c}{Subway} \\
 & PSNR-Y & FLIP & JOD & PSNR-Y & FLIP & JOD & PSNR-Y & FLIP & JOD & PSNR-Y & FLIP & JOD \\
\midrule
Bicubic & 33.24 & 0.0561 & 8.93 & 32.83 & 0.0665 & 8.25 & 31.42 & 0.0804 & 7.93 & 29.87 & 0.0700 & 7.36 \\
TAAU & 32.18 & 0.0714 & 8.83 & 32.90 & 0.0698 & 8.31 & 31.12 & 0.0862 & 7.86 & 30.39 & 0.0712 & 7.54 \\
TAAU, clamp & 32.11 & 0.0724 & 8.81 & 32.81 & 0.0704 & 8.29 & 31.02 & 0.0873 & 7.83 & 30.37 & 0.0719 & 7.54 \\
Ours, $\lambda_T{=}0$ & 34.16 & 0.0710 & 8.99 & 37.53 & 0.0466 & 9.16 & 31.64 & 0.0866 & 7.98 & 31.08 & 0.0703 & 7.65 \\
Ours, $\lambda_T{=}0.05$ & 33.61 & 0.0875 & 8.86 & 37.62 & 0.0476 & 9.17 & 31.61 & 0.0865 & 7.93 & 31.09 & 0.0710 & 7.68 \\
Ours, $\lambda_T{=}0.25$ & 30.46 & 0.1379 & 8.25 & 37.49 & 0.0505 & 9.13 & 29.35 & 0.0999 & 7.08 & 29.13 & 0.0840 & 6.77 \\
\bottomrule
\end{tabular}
\par\smallskip
\begin{minipage}{\linewidth}\scriptsize
All frames of each test range at 60\,Hz, reset per sequence, 8 warm-up frames; FLIP LDR 67\,ppd; ColorVideoVDP 0.5.7. Our STSS models are trained on the first Lewis training file only, with 16-frame clips (30 in our QRISP recipe).
\end{minipage}
\end{table*}
%% ======== END tables/T7s_stss_ours ========

\section{Ablations}\label{supp:section-8}
We separate the recipe ablations from the registration ablations in Table~\ref{supp:tab:ablation}
(Section~\ref{sec:ablations} of the paper).

At 45k iterations, the normalised accumulator lowers static variation at comparable PSNR to the single-$\alpha$
blend. The two rows have the same displayed temporal PSNR, but different static standard deviations: the
temporal metric averages warped frame differences, whereas the static statistic measures variation during the
camera stop. Using 16-frame clips lowers the static standard deviation further and costs 0.14\,dB in PSNR.

The registration block removes one component at a time with no temporal loss, using 15k training runs.
The successful component removals cost 0.03 to 0.14\,dB relative to the control with the same seed.
The two controls differ by 0.06\,dB. Removing unjittered depth makes seed 0 non-finite from about 11k of
15k training iterations, so we report seed 1 against its own control and retain the failed run in
Table~\ref{supp:tab:loo-full}.

%% ======== BEGIN tables/T4_ablations ========
% T4_ablations.tex -- generated by paper/make_tables.py (2026-10-01 15:56 UTC); do not edit by hand
% except the caption, which is preserved when the table is regenerated.
% Every number is read from the file named in the '% src:' comment above its row.
\begin{table}[t]
\centering
\caption{Ablations. (a) Components of the final recipe at 45k iterations. (b) One registration component removed
at a time, with $\lambda_T=0$ and 15k iterations. Each difference uses the control with the same seed;
the failed seed-0 depth run is retained in Table~\ref{supp:tab:loo-full}.}
\label{supp:tab:ablation}
\footnotesize
\setlength{\tabcolsep}{2.5pt}
\begin{tabular}{@{}l r r r r r@{}}
\toprule
Variant & PSNR-Y$\uparrow$ & $\Delta$ & LPIPS$\downarrow$ & tPSNR$_{60}\!\uparrow$ & Std$_{\mathrm{s}}$ \\
\midrule
\multicolumn{6}{@{}l}{\textit{(a) R, $\lambda_T=0.05$, 192$\times$8, $\times$4, 45k}} \\
final recipe & 29.74 & --- & 0.196 & 30.95 & 0.43 \\
%      (difference of the two numbers above)
single-$\alpha$ blend accumulator & 29.71 & $-$0.02 & 0.195 & 30.95 & 0.53 \\
%      (difference of the two numbers above)
16-frame training clips & 29.60 & $-$0.14 & 0.208 & 30.95 & 0.33 \\
\midrule
\multicolumn{6}{@{}l}{\textit{(b) R minus one piece, $\lambda_T=0$, 192$\times$8, $\times$4, 15k}} \\
control, all pieces & 29.31 & --- & 0.203 & 30.19 & 0.83 \\
control, all pieces, seed 1 & 29.38 & --- & 0.223 & 30.25 & 0.74 \\
%      (difference of the two numbers above)
$-$ sample phase & 29.22 & $-$0.09 & 0.208 & 30.08 & 0.81 \\
%      (difference of the two numbers above)
$-$ registered skip & 29.28 & $-$0.03 & 0.205 & 30.14 & 0.80 \\
%      (difference of the two numbers above)
$-$ LR re-jitter & 29.19 & $-$0.12 & 0.210 & 30.12 & 0.81 \\
%      (difference of the two numbers above)
$-$ unjittered depth, seed 1 & 29.23 & $-$0.14 & 0.209 & 30.08 & 0.76 \\
\bottomrule
\end{tabular}
\par\smallskip
\begin{minipage}{\linewidth}\scriptsize
$\Delta$: PSNR-Y difference to the control of the same block and seed (seed 0 unless noted). (a) final recipe = normalised accumulator and 30-frame training clips. (b) the registration pieces of Section~\ref{sec:registration} of the paper; Table~\ref{supp:tab:loo-full} lists the remaining leave-one-out runs. Columns: PSNR-Y and its difference $\Delta$, LPIPS, tPSNR$_{60}$ and Std$_{\mathrm s}$, defined as in Table~\ref{supp:tab:grid}.
\end{minipage}
\end{table}
%% ======== END tables/T4_ablations ========

\section{Registration leave-one-out, all runs}\label{supp:section-9}
Table~\ref{supp:tab:loo-full} lists every leave-one-out run behind Section~\ref{sec:ablations} of the paper, including the
float32 accumulator, a variant that keeps only the unjittered motion field, and the seed-0 run without unjittered
depth, which diverged in training (non-finite from about 11k of 15k iterations) and was rerun with seed 1.

Across the successfully scored leave-one-out runs, PSNR drops by 0.01 to 0.14\,dB. The float32-accumulator
row is included in this range, and the variant retaining only unjittered motion removes several registration
components together. We therefore distinguish it from the rows that remove a single component.

These zero-weight runs measure reconstruction accuracy and training failure after a component is removed.
The registration-by-temporal-weight experiments in Section~\ref{sec:mechanism} of the paper test the
effect on the temporal loss. A small PSNR difference at zero weight does not measure whether the removed
component changes copying at nonzero weight.

%% ======== BEGIN tables/T4s_loo_full ========
% T4s_loo_full.tex -- generated by paper/make_tables.py (2026-10-01 15:56 UTC); do not edit by hand
% except the caption, which is preserved when the table is regenerated.
% Every number is read from the file named in the '% src:' comment above its row.
\begin{table}[t]
\centering
\caption{All registration leave-one-out runs: R, $\lambda_T=0$, $192\times8$, $\times$4, 15k iterations.
This includes the failed seed-0 depth run and complements Table~\ref{supp:tab:ablation}.}
\label{supp:tab:loo-full}
\footnotesize
\setlength{\tabcolsep}{2.5pt}
\begin{tabular}{@{}l r r r r r@{}}
\toprule
Variant & PSNR-Y$\uparrow$ & $\Delta$ & LPIPS$\downarrow$ & tPSNR$_{60}\!\uparrow$ & Std$_{\mathrm{s}}$ \\
\midrule
control, all pieces & 29.31 & --- & 0.203 & 30.19 & 0.83 \\
control, all pieces, seed 1 & 29.38 & --- & 0.223 & 30.25 & 0.74 \\
%      (difference of the two numbers above)
$-$ sample phase & 29.22 & $-$0.09 & 0.208 & 30.08 & 0.81 \\
%      (difference of the two numbers above)
$-$ registered skip & 29.28 & $-$0.03 & 0.205 & 30.14 & 0.80 \\
%      (difference of the two numbers above)
$-$ LR re-jitter & 29.19 & $-$0.12 & 0.210 & 30.12 & 0.81 \\
% b5_lo_depth: 1500 non-finite frames in the scored pass -> 'div.'
$-$ unjittered depth & div. & div. & div. & div. & div. \\
%      (difference of the two numbers above)
$-$ unjittered depth, seed 1 & 29.23 & $-$0.14 & 0.209 & 30.08 & 0.76 \\
%      (difference of the two numbers above)
$-$ fp32 accumulator & 29.30 & $-$0.01 & 0.204 & 30.17 & 0.83 \\
%      (difference of the two numbers above)
only unjittered motion & 29.21 & $-$0.10 & 0.212 & 30.02 & 0.82 \\
\bottomrule
\end{tabular}
\par\smallskip
\begin{minipage}{\linewidth}\scriptsize
$\Delta$: PSNR-Y difference to the control with the same seed (seed 0 unless noted). `only unjittered motion' keeps the non-jittered motion field and removes every other piece. div.: the seed-0 run with jittered depth diverged in training (non-finite from about 11k of 15k iterations; repeated with seed 1). Columns: PSNR-Y and its difference $\Delta$, LPIPS, tPSNR$_{60}$ and Std$_{\mathrm s}$, defined as in Table~\ref{supp:tab:grid}.
\end{minipage}
\end{table}
%% ======== END tables/T4s_loo_full ========

\section{Results at $\times$3 and $\times$4}\label{supp:section-10}
\begin{figure*}[!tbp]
  \centering
  \includegraphics[width=\textwidth]{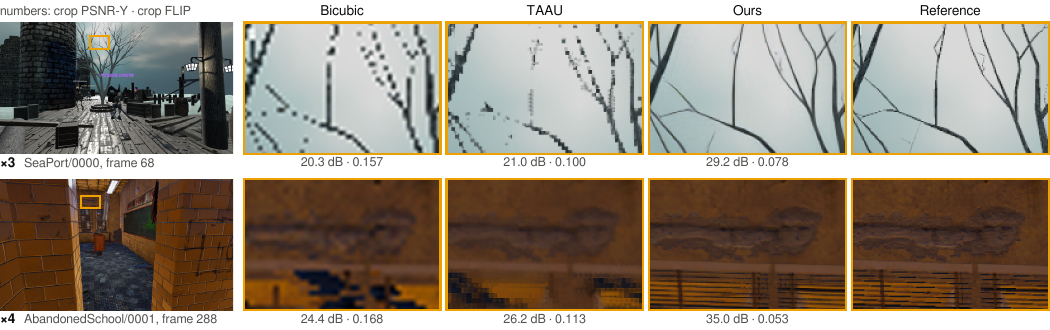}
  \caption{Selected crops at $\times$3 and $\times$4: bicubic, our classical TAAU, our model and the reference.
The full reference frames mark the crop locations, and labels give crop PSNR-Y and FLIP.
The $\times$4 crop lies in the static segment. Selection is described in the text and in
Section~\ref{supp:sec:gallery}.}
  \label{supp:fig:scales}
\end{figure*}

Figure~\ref{supp:fig:scales} shows the crop with the largest gain over the best baseline at $\times$3 and $\times$4; the
gallery at the end of this document shows two crops for every scene and scale.

For the two displayed scales, the scene assignment maximises the summed PSNR-Y gain. Within each assigned
scene, we select the largest gain over the best baseline among windows satisfying the detail criterion in
Section~\ref{supp:sec:gallery} and having lower FLIP than every baseline. The $\times$4 crop is in the static
segment. These are selected detail examples; the scene and sequence averages are reported separately.

The thin branches at $\times$3 and the railing at $\times$4 expose different spatial errors: the baseline
outputs break these structures into dots or blocks, while our output follows the reference more closely.
The gallery uses slightly different windows on the same frames, so its crop scores differ.
The labels report each window's score; the full-sequence results use all scored frames.

\section{Latency with matched engineering}\label{supp:section-11}
\begin{figure}[!htbp]
  \centering
  \includegraphics[width=\columnwidth]{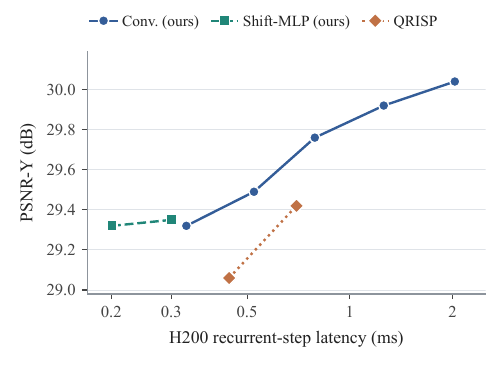}
  \caption{PSNR-Y and H200 recurrent-step latency at $\times$4,
  from Table~\ref{supp:tab:latency}. Circles: our convolutional trunks;
  squares: our shift-MLP trunks, both trained for 200k iterations.
  Diamonds: published QRISP PSNR paired with our
  re-implementation's latency. Lines connect configurations
  within each method.}
  \label{supp:fig:pareto}
\end{figure}

Table~\ref{supp:tab:latency} repeats the current H200 timings of Table~\ref{tab:headline} of the paper and adds
deployed parameter and GFLOP counts. Figure~\ref{supp:fig:pareto} uses only these current measurements.
Table~\ref{supp:tab:int8} measures quantisation error, not inference time.

The current timing measures input preparation, reprojections, the network, accumulation and output formation
at $\times$4, 1080p output and batch 1, excluding rendering. Every timed configuration uses fused kernels,
bfloat16 with float32 accumulation, and one CUDA graph. We average 300 calls after 50 warm-up calls.
QRISP's PSNR is published quality; its time is measured on our random-weight re-implementation with the
same engineering. These are architecture comparisons under the stated harness, with different training budgets.

Our $96\times4$ network reaches 29.49\,dB at 0.524\,ms, compared with QRISP-L's 29.42\,dB at 0.698\,ms:
0.07\,dB higher PSNR at 25\% lower latency. The $64\times6$ and $64\times8$ shift-MLP variants reach 29.32
and 29.35\,dB at 0.201 and 0.301\,ms. They are 0.10 and 0.07\,dB below QRISP-L at 71\% and 57\% lower
latency, and 0.26 and 0.29\,dB above QRISP-M's 29.06\,dB at 0.443\,ms. The largest model gains 0.62\,dB
over QRISP-L at 2.032\,ms, or 2.91 times its measured latency.

The int8 results measure the complete recurrent rollout, including the effect of earlier quantised outputs on
later inputs. Calibration frames come from the scored sequences, as specified in Table~\ref{supp:tab:int8}; the
reported accuracy cost belongs to that calibration set and supplies no timing comparison.

%% ======== BEGIN tables/T5_latency ========
% All H200 rows are synchronised with main Table 2.
\begin{table*}[t]
\centering
\caption{Recurrent-step latency at $\times$4, 1080p output and batch 1, excluding rendering.
H200 measurements are identical to the main paper. Daggered PSNR values are published QRISP results.}
\label{supp:tab:latency}
\footnotesize
\setlength{\tabcolsep}{3pt}
\begin{tabular*}{\textwidth}{@{\extracolsep{\fill}}l l r r r r r@{}}
\toprule
Model & Size & Iter. & Params & GFLOP & H200 ms & PSNR-Y$\,\uparrow$ \\
\midrule
\multicolumn{7}{@{}l}{\textit{Measurements from main Table~\ref{tab:headline}}} \\
QRISP-M$^{\dagger}$ & f32 / m3 & 500k & 71k & 18 & 0.443 & 29.06 \\
QRISP-L$^{\dagger}$ & f64 / m5 & 500k & 0.27M & 70 & 0.698 & 29.42 \\
Ours, shift-MLP & 64$\times$6 & 200k & 0.17M & 38 & 0.201 & 29.32 \\
Ours, shift-MLP & 64$\times$8 & 200k & 0.20M & 47 & 0.301 & 29.35 \\
Ours & 64$\times$4 & 200k & 0.36M & 89 & 0.332 & 29.32 \\
Ours & 96$\times$4 & 200k & 0.77M & 191 & 0.524 & 29.49 \\
Ours & 96$\times$8 & 200k & 1.43M & 363 & 0.790 & 29.76 \\
Ours & 192$\times$8 & 200k & 5.51M & 1414 & 1.258 & 29.92 \\
Ours & 256$\times$8 & 200k & 9.70M & 2496 & 2.032 & 30.04 \\
\bottomrule
\end{tabular*}
\par\smallskip
\begin{minipage}{\linewidth}\scriptsize
Fused Triton convolutions and pre/post kernels, one CUDA graph, bf16 with fp32 accumulation;
300 calls after 50 warm-ups.
$^{\dagger}$QRISP~\cite{mercier2023efficient}: published PSNR at 500k iterations, paired with our random-weight
re-implementation's time.
The $64\times4$ to $96\times8$ convolutional models use re-parameterised blocks folded for inference.
Params: deployed parameters. GFLOP: twice the multiply-adds per recurrent step; the elementwise output stage
is excluded from this arithmetic count but included in latency.
QRISP f$X$/m$Y$: $X$ features and $Y$ intermediate reconstruction layers; ours: channels $\times$ blocks.
\end{minipage}
\end{table*}
%% ======== END tables/T5_latency ========

%% ======== BEGIN tables/T8s_int8 ========
% T8s_int8.tex -- generated by paper/make_tables.py (2026-10-01 15:56 UTC); do not edit by hand
% except the caption, which is preserved when the table is regenerated.
% Every number is read from the file named in the '% src:' comment above its row.
\begin{table}[t]
\centering
\caption{PSNR cost of TensorRT int8 inference ($\times$4): change in PSNR-Y against the float32 model over one unbroken 300-frame pass of each scene.
}
\label{supp:tab:int8}
\footnotesize
\setlength{\tabcolsep}{4pt}
\begin{tabular}{@{}l l r r r r@{}}
\toprule
 & & & \multicolumn{3}{c}{$\Delta$PSNR-Y (dB)} \\
\cmidrule(l){4-6}
Model & GPU & Iter. & fp16 & int8 & int8, worst scene \\
\midrule
64$\times$4 & A100 & 15k & 0.000 & $-$0.128 & $-$0.214 \\
64$\times$4 & H200 & 15k & 0.000 & $-$0.127 & $-$0.214 \\
192$\times$8 & A100 & 45k & 0.000 & $-$0.352 & $-$0.489 \\
192$\times$8 & H200 & 45k & 0.000 & $-$0.341 & $-$0.480 \\
\bottomrule
\end{tabular}
\par\smallskip
\begin{minipage}{\linewidth}\scriptsize
TensorRT~10.16 engines of one recurrent step. int8: int8 and fp16 in the trunk (stem, blocks, heads and projections) with entropy calibration on 50 steady-state inputs in total (10 consecutive frames of each of the five test sequences); fp16: fp16 trunk. The 1080p stage stays in float32 in both. $\Delta$PSNR-Y against the eager float32 model run in lockstep on AbandonedSchool/0000, SeaPort/0000 and SpaceShipDemo/0000: mean over the three passes and the worst pass. 64$\times$4: plain control; 192$\times$8: the analysis architecture. The calibration frames lie inside the scored passes, so these results measure quantisation error on sequences that also supply calibration inputs.
\end{minipage}
\end{table}
%% ======== END tables/T8s_int8 ========

\section{Run time in a frame budget}\label{supp:section-12}\label{supp:sec:budget}
Table~\ref{supp:tab:budget} converts the current H200 timings of Table~\ref{supp:tab:latency} into shares of
60\,Hz and 120\,Hz frame intervals. The shift-MLP trunks and $64\times4$ convolutional model take about
1\% to 2\% of a 60\,Hz frame and 2\% to 4\% of a 120\,Hz frame. The most accurate model takes 12\%
and 24\%, respectively. These percentages account only for upscaling; rendering and other work consume
the remaining budget.

At $\times$4 the renderer shades one pixel in sixteen. If rendering cost scales with shaded-pixel count,
the break-even native rates are 461\,fps for the $256\times8$ model and 2\,824\,fps for the $64\times4$ model.
For the six-block shift-MLP the rate is 4\,664\,fps. These are thresholds under the pixel-cost model,
not measured game frame rates: geometry, shadow maps and other resolution-independent costs reduce the saving.
Registration changes fields and offsets in existing warps, and the temporal weight changes training;
neither adds inference operations to a fixed architecture.

\paragraph*{The recipe on QRISP's architecture.} We also apply the recipe to our QRISP-L re-implementation with
registered inputs (Table~\ref{supp:tab:transfer}). The temporal weight changes training but adds no inference operations. At 45k iterations, averaged over two seeds, the small weight of our recipe nearly
halves the static pixel standard deviation (1.14 to 0.61) at 0.06\,dB lower PSNR and 0.17\,dB lower static PSNR.
At 15k iterations the large weight freezes the accumulator (median $\alpha$ $1.1\cdot10^{-5}$) and costs 0.93\,dB of static PSNR.

The QRISP-L transfer runs separate the inference architecture from the training objective. Their timing is
unchanged when the temporal weight changes, while their static variation and accuracy change.
The small weight at 45k iterations reduces static variation with a smaller accuracy cost than the large weight
in the 15k comparison. These are within-protocol observations; the published QRISP score uses a different
pipeline and training length.

%% ======== BEGIN tables/T9s_budget ========
% Derived only from the current H200 timings in main Table 2.
\begin{table}[t]
\centering
\caption{Current H200 run time as a share of a frame interval ($\times$4, 1080p output).
Percentages and break-even thresholds are derived from Table~\ref{supp:tab:latency}, with exact intervals
$1000/60$ and $1000/120$\,ms.}
\label{supp:tab:budget}
\footnotesize
\setlength{\tabcolsep}{2.3pt}
\begin{tabular*}{\columnwidth}{@{\extracolsep{\fill}}l r r r r@{}}
\toprule
Model & H200 ms & 60\,Hz & 120\,Hz & Break-even$^{\ast}$ \\
\midrule
QRISP-M & 0.443 & 3\% & 5\% & 2\,116\,fps \\
QRISP-L & 0.698 & 4\% & 8\% & 1\,343\,fps \\
Ours, shift-MLP $64\times6$ & 0.201 & 1\% & 2\% & 4\,664\,fps \\
Ours, shift-MLP $64\times8$ & 0.301 & 2\% & 4\% & 3\,115\,fps \\
Ours $64\times4$ & 0.332 & 2\% & 4\% & 2\,824\,fps \\
Ours $96\times4$ & 0.524 & 3\% & 6\% & 1\,789\,fps \\
Ours $96\times8$ & 0.790 & 5\% & 9\% & 1\,187\,fps \\
Ours $192\times8$ & 1.258 & 8\% & 15\% & 745\,fps \\
Ours $256\times8$ & 2.032 & 12\% & 24\% & 461\,fps \\
\bottomrule
\end{tabular*}
\par\smallskip
\begin{minipage}{\linewidth}\scriptsize
Percentages are rounded to whole percent; frame rates to the nearest fps.
$^{\ast}$At $\times$4, pixel-scaled rendering costs $1/16$ of native rendering. For upscaling time $t$\,ms,
$T_{\mathrm{native}}/16+t<T_{\mathrm{native}}$ when $T_{\mathrm{native}}>16t/15$; the corresponding native rate is below
$1000/(16t/15)$\,fps. This calculation excludes costs that do not scale with resolution and does not predict
the complete rendered frame rate.
\end{minipage}
\end{table}
%% ======== END tables/T9s_budget ========

%% ======== BEGIN tables/T9s_transfer ========
% T9s_transfer.tex -- generated by paper/make_tables.py (2026-10-01 15:56 UTC); do not edit by hand
% except the caption, which is preserved when the table is regenerated.
% Every number is read from the file named in the '% src:' comment above its row.
\begin{table}[t]
\centering
\caption{The QRISP-style network of Section~\ref{sec:generality} of the paper (QRISP-L's architecture) trained with our registered pipeline. A small temporal weight nearly halves the static pixel standard deviation at 0.06\,dB lower PSNR (45k iterations, two seeds); a large one freezes the accumulator and costs 0.9\,dB of static PSNR (15k). The scores are not comparable with the published ones (500k iterations, their pipeline).}
\label{supp:tab:transfer}
\footnotesize
\setlength{\tabcolsep}{2pt}
\begin{tabular*}{\columnwidth}{@{\extracolsep{\fill}}l l l r r r r r@{}}
\toprule
Iter. & $\lambda_T$ & Seed & PSNR-Y$\,\uparrow$ & LPIPS$\,\downarrow$ & Static PSNR$\,\uparrow$ & Static std & Med.\ $\alpha$ \\
\midrule
15k & 0 & 0 & 28.73 & 0.241 & 33.59 & 1.30 & 0.031 \\
15k & 0.05 & 0 & 28.69 & 0.261 & 33.53 & 0.64 & 1.2$\cdot 10^{-3}$ \\
15k & 0.25 & 0 & 28.41 & 0.313 & 32.66 & 0.23 & 1.1$\cdot 10^{-5}$ \\
\midrule
45k & 0 & 0 & 29.05 & 0.210 & 33.89 & 1.16 & 0.024 \\
45k & 0 & 1 & 29.04 & 0.213 & 33.89 & 1.12 & 0.024 \\
45k & 0.05 & 0 & 28.97 & 0.233 & 33.71 & 0.59 & 6.9$\cdot 10^{-4}$ \\
45k & 0.05 & 1 & 29.00 & 0.232 & 33.73 & 0.64 & 1.7$\cdot 10^{-3}$ \\
\bottomrule
\end{tabular*}
\par\smallskip
\begin{minipage}{\linewidth}\scriptsize
Our re-implementation of QRISP-L (f64\,/\,m5; the architecture profiled at 0.698\,ms on the H200 in Table~\ref{supp:tab:latency}; the temporal weight changes training only), $\times$4, registered inputs, our reconstruction loss and schedule. Static PSNR and std computed after quantising the output to 8-bit values, on AbandonedSchool/0001 frames 275--292; median $\alpha$ over that segment.
\end{minipage}
\end{table}
%% ======== END tables/T9s_transfer ========

\section{Trunk design and learning rate}\label{supp:section-13}
Table~\ref{supp:tab:design} compares trunk alternatives at $\times$4 after 15k iterations. All use the registered
pipeline and recipe; the table specifies the trunk, width, depth and learning rate. PSNR differences use the
first control of the same tier, except where a second seed has its own control. Deployment latency is
reported separately in Table~\ref{supp:tab:latency}, for the configurations in the main paper.

\paragraph*{Conventional trunks (a).}
The tested U-Nets range from a 0.01\,dB gain to a 0.27\,dB loss against their tier's plain-convolution control.
Partial convolutions, star blocks and inverted
residual blocks with depthwise convolutions lose 0.29 to 0.56\,dB at the wide-network learning rate.
In the small tier, re-parameterised convolution blocks gain 0.24\,dB at the control learning rate, but a plain
network gains 0.23\,dB by raising its learning rate; combining the two changes gives only 0.13\,dB.
In the medium tier, the ECB gains are 0.18 and 0.21\,dB across two seeds.
These results make learning rate part of the architecture comparison.

\paragraph*{Light token mixers (b).}
The six-block, 64-channel shift-MLP reaches 28.90\,dB, 0.08\,dB below the plain-convolution control.
The tested depthwise, pooling, partial-convolution and linear-attention mixers all score lower.
For the four- and six-block 64-channel shift-MLPs, removing recurrent features and replacing the first
convolution by a jitter-conditioned $1\times1$ convolution lowers PSNR by 0.13 and 0.14\,dB, respectively.
These are short-training accuracy comparisons; the 200k models and their measured latencies are reported
in Table~\ref{supp:tab:latency}.

%% ======== BEGIN tables/T6_design ========
% Accuracy values retained from the supplied trunk design study; timing
% columns and all derived speed comparisons have been removed.
\begin{table*}[t]
\centering
\caption{Trunk alternatives and learning rates: $\times$4, 15k iterations, registered pipeline and the same
training recipe. Differences use the first control of the same tier, with the medium-tier seed-1 ECB
compared against its seed-1 control. Only reconstruction accuracy is reported.}
\label{supp:tab:design}
\footnotesize
\setlength{\tabcolsep}{4pt}
\begin{tabular}{@{}l l l r r r@{}}
\toprule
Tier & Trunk & Size & lr & PSNR-Y$\,\uparrow$ & $\Delta$PSNR \\
\midrule
\multicolumn{6}{@{}l}{\textit{(a) Conventional trunks}} \\
\midrule
S & plain conv (control) & 64$\times$4 & 6.0e-4 & 28.75 & --- \\
 & ECB re-parameterised & 64$\times$4 & 6.0e-4 & 28.98 & +0.24 \\
 & plain conv & 64$\times$4 & 2.0e-3 & 28.98 & +0.23 \\
 & ECB re-parameterised & 64$\times$4 & 2.0e-3 & 28.88 & +0.13 \\
 & U-Net 1-2-1 & 64/128 & 6.0e-4 & 28.76 & +0.01 \\
 & U-Net 1-2-1 & 48/96 & 6.0e-4 & 28.62 & $-$0.13 \\
\midrule
M & plain conv (control) & 96$\times$8 & 6.0e-4 & 29.16 & --- \\
 & plain conv (control, seed 1) & 96$\times$8 & 6.0e-4 & 29.17 & 0.00 \\
 & ECB re-parameterised & 96$\times$8 & 6.0e-4 & 29.37 & +0.21 \\
 & ECB re-parameterised (seed 1) & 96$\times$8 & 6.0e-4 & 29.35 & +0.18 \\
 & U-Net 1-7-1 & 96/192 & 6.0e-4 & 29.10 & $-$0.06 \\
 & U-Net 1-7-1, ECB & 96/192 & 6.0e-4 & 28.90 & $-$0.27 \\
 & U-Net, 2 levels & 96/192/384 & 6.0e-4 & 28.89 & $-$0.27 \\
 & StarNet & 96$\times$8 & 6.0e-4 & 28.83 & $-$0.33 \\
 & FasterNet (PConv) & 128$\times$6 & 6.0e-4 & 28.87 & $-$0.29 \\
 & MobileNetV2 & 64$\times$8 & 6.0e-4 & 28.60 & $-$0.56 \\
\midrule
L & plain conv (control) & 192$\times$8 & 6.0e-4 & 29.48 & --- \\
 & U-Net 1-7-1 & 192/384 & 6.0e-4 & 29.32 & $-$0.16 \\
\midrule
\multicolumn{6}{@{}l}{\textit{(b) Light token mixers}} \\
\midrule
Light & plain conv (control) & 64$\times$4 & 2.0e-3 & 28.98 & --- \\
 & plain conv, fd 0 & 64$\times$4 & 2.0e-3 & 28.89 & $-$0.09 \\
 & plain conv, fast & 64$\times$4 & 2.0e-3 & 28.91 & $-$0.06 \\
 & shift-MLP & 48$\times$4 & 2.0e-3 & 28.67 & $-$0.30 \\
 & shift-MLP & 48$\times$8 & 2.0e-3 & 28.84 & $-$0.14 \\
 & shift-MLP, ReLU & 48$\times$6 & 2.0e-3 & 28.74 & $-$0.24 \\
 & shift-MLP & 64$\times$4 & 2.0e-3 & 28.76 & $-$0.21 \\
 & shift-MLP & 64$\times$4 & 4.0e-3 & 28.80 & $-$0.18 \\
 & shift-MLP & 64$\times$6 & 2.0e-3 & 28.90 & $-$0.08 \\
 & shift-MLP, fast & 48$\times$3 & 2.0e-3 & 28.50 & $-$0.48 \\
 & shift-MLP, fast & 64$\times$3 & 2.0e-3 & 28.59 & $-$0.39 \\
 & shift-MLP, fast & 48$\times$4 & 2.0e-3 & 28.54 & $-$0.44 \\
 & shift-MLP, fast & 64$\times$4 & 2.0e-3 & 28.63 & $-$0.35 \\
 & shift-MLP, fast, ReLU & 64$\times$4 & 2.0e-3 & 28.64 & $-$0.34 \\
 & shift-MLP, fast & 64$\times$6 & 2.0e-3 & 28.76 & $-$0.21 \\
 & shift-MLP, fast & 48$\times$8 & 2.0e-3 & 28.71 & $-$0.27 \\
 & depthwise MLP & 48$\times$4 & 2.0e-3 & 28.61 & $-$0.37 \\
 & depthwise MLP & 64$\times$4 & 2.0e-3 & 28.72 & $-$0.25 \\
 & PConv-GLU & 48$\times$4 & 2.0e-3 & 28.57 & $-$0.41 \\
 & pool-MLP & 48$\times$4 & 2.0e-3 & 28.68 & $-$0.29 \\
 & linear attention & 48$\times$4 & 2.0e-3 & 28.65 & $-$0.33 \\
\bottomrule
\end{tabular}
\par\smallskip
\begin{minipage}{\linewidth}\scriptsize
$\Delta$PSNR uses the first control of the same tier, except that seed-1 rows use the seed-1 control. ECB: every $3{\times}3$ convolution is trained as the multi-branch edge-oriented convolution block of ECBSR~\cite{zhang2021edge} and folded into a single $3{\times}3$ convolution for inference. U-Net $a$-$b$-$c$: $a$ blocks at input resolution, $b$ blocks at half resolution with twice the channels, then $c$ blocks at input resolution after a skip connection (Size: channels per level); 2 levels: 1-1-4-1-1 blocks over three resolutions. `fd 0': no recurrent feature channels; `fast': fd 0 and a 1$\times$1 instead of a 3$\times$3 jitter-conditioned first convolution.
\end{minipage}
\end{table*}
%% ======== END tables/T6_design ========

\section{Failure case: particles without motion vectors}\label{supp:section-14}
\begin{figure}[!htbp]
  \centering
  \includegraphics[width=\columnwidth]{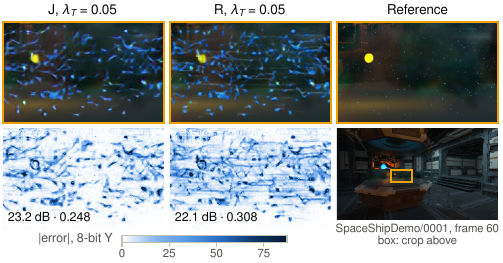}
  \caption{Failure case: particle sprites without motion vectors (SpaceShipDemo/0001, $\times$4; the R and J grid models at $\lambda_T=0.05$, Table~\ref{supp:tab:scenes} gives both weights). Both models smear the particle sprites. The R model has larger error in the shown crop (22.1 versus 23.2\,dB crop PSNR-Y; the labels give crop PSNR-Y and \reflectbox{F}LIP).}
  \label{supp:fig:failure}
\end{figure}

Figure~\ref{supp:fig:failure} shows the SpaceShipDemo sequence discussed in Section~\ref{sec:limitations} of the paper, and
Table~\ref{supp:tab:scenes} reports the per-scene comparison. At $\lambda_T=0$, R scores 0.13\,dB lower than J on
SpaceShipDemo, with the deficit confined to sequence 0001; R leads on the other two scenes. Figure~\ref{supp:fig:failure}
illustrates the particle failure at $\lambda_T=0.05$, where R scores lower on both SpaceShipDemo sequences.

Registration can align a history only where the supplied motion field describes the content being transported.
The particle sprites lack their own motion vectors, so correcting the lattice conventions does not resolve
their alignment. Both outputs show errors in the displayed crop. At zero temporal weight, the per-sequence columns locate the R model's deficit in the sequence containing
these sprites. At the small weight, the displayed crop also has larger error under R.

\section{Additional measurements}\label{supp:section-15}
\label{supp:sec:extra}
We report the stateless, history, per-sequence and seed comparisons used in the paper, retaining the
metric and aggregation conventions of each experiment.

\paragraph*{Stateless upsamplers and networks without history.}
Table~\ref{supp:tab:extra-rows}a scores Lanczos-3 upsampling on the unjittered input of the bicubic rows of
Table~\ref{tab:headline} of the paper: it is 0.14\,dB below bicubic upsampling at every scale. On the jittered inputs used by the networks, Lanczos-3 scores
27.95 and 25.17\,dB at $\times$2 and $\times$4, respectively; these are 1.36 and 0.98\,dB below bicubic on the
unjittered inputs.
Table~\ref{supp:tab:extra-rows}b compares our recurrent network with the same network without history, both trained for
45k iterations: without history it loses 1.65\,dB at $\times$2 and 2.02\,dB at $\times$4 on QRISP and 4.43\,dB
zero-shot on the Arm capture, and its static pixel standard deviation rises from 0.11 to 1.09 ($\times$2) and from
0.44 to 1.74 ($\times$4).

The stateless rows also show why temporal standard deviation should accompany an accuracy measurement.
Bicubic and Lanczos have very low static variation, but their spatial scores are below those of the recurrent
network. Removing history from the network lowers PSNR and raises static variation at both tested scales.
We use PSNR together with static variation to assess the effect of history.

%% ======== BEGIN tables/T8s_extra_rows ========
% T8s_extra_rows.tex -- generated by paper/make_tables.py (2026-10-01 15:56 UTC); do not edit by hand
% except the caption, which is preserved when the table is regenerated.
% Every number is read from the file named in the '% src:' comment above its row.
\begin{table*}[t]
\centering
\caption{Additional rows. (a) Stateless and classical temporal upsamplers on QRISP ($\Delta$: against bicubic upsampling; for the clamp rows, against the TAAU with variance clipping). (b) Our network with and without history at equal training length (45k iterations), on QRISP and zero-shot on the Arm capture.
}
\label{supp:tab:extra-rows}
\footnotesize
\setlength{\tabcolsep}{4pt}
\begin{tabular*}{\textwidth}{@{\extracolsep{\fill}}l r r r r r r r@{}}
\toprule
Method & PSNR-Y$\,\uparrow$ & $\Delta$ & SSIM-Y$\,\uparrow$ & LPIPS$\,\downarrow$ & FLIP$\,\downarrow$ & JOD$\,\uparrow$ & Static std \\
\midrule
\multicolumn{8}{@{}l}{\textit{(a) Stateless and classical temporal upsamplers, QRISP}} \\
\multicolumn{8}{@{}l}{$\times$2} \\
\quad Bicubic & 29.31 &  & 0.8649 & 0.217 & 0.0719 & 7.81 & 0.011 \\
%      (difference of the two numbers above)
\quad Lanczos-3 & 29.17 & $-$0.14 & 0.8622 & 0.233 & 0.0718 & 7.78 & 0.011 \\
%      (difference of the two numbers above)
\quad Lanczos-3, jittered input & 27.95 & $-$1.36 & 0.8271 & 0.237 & --- & --- & --- \\
\quad TAAU (variance clipping) & 29.99 &  & 0.8740 & 0.162 & 0.0743 & 7.99 & 0.67 \\
%      (difference of the two numbers above)
\quad TAAU, neighbourhood clamping & 29.93 & $-$0.06 & 0.8741 & 0.161 & 0.0750 & 7.96 & 0.66 \\
\multicolumn{8}{@{}l}{$\times$3} \\
\quad Bicubic & 27.37 &  & 0.8009 & 0.370 & 0.0999 & 6.83 & 0.0090 \\
%      (difference of the two numbers above)
\quad Lanczos-3 & 27.23 & $-$0.14 & 0.7961 & 0.378 & 0.1013 & 6.77 & 0.0098 \\
\quad TAAU (variance clipping) & 28.59 &  & 0.8299 & 0.233 & 0.0876 & 7.36 & 0.86 \\
%      (difference of the two numbers above)
\quad TAAU, neighbourhood clamping & 28.59 & $-$0.01 & 0.8315 & 0.231 & 0.0876 & 7.36 & 0.85 \\
\multicolumn{8}{@{}l}{$\times$4} \\
\quad Bicubic & 26.15 &  & 0.7534 & 0.444 & 0.1225 & 6.11 & 0.0090 \\
%      (difference of the two numbers above)
\quad Lanczos-3 & 26.00 & $-$0.14 & 0.7479 & 0.450 & 0.1251 & 6.04 & 0.0099 \\
%      (difference of the two numbers above)
\quad Lanczos-3, jittered input & 25.17 & $-$0.98 & 0.7171 & 0.453 & --- & --- & --- \\
\quad TAAU (variance clipping) & 27.61 &  & 0.7938 & 0.295 & 0.0997 & 6.82 & 1.06 \\
%      (difference of the two numbers above)
\quad TAAU, neighbourhood clamping & 27.63 & +0.02 & 0.7962 & 0.291 & 0.0992 & 6.85 & 1.05 \\
\midrule
\multicolumn{8}{@{}l}{\textit{(b) Our network (192$\times$8), 45k iterations}} \\
\multicolumn{8}{@{}l}{QRISP, $\times$2} \\
\quad recurrent & 32.58 &  & 0.9220 & 0.095 & 0.0579 & 8.60 & 0.11 \\
%      (difference of the two numbers above)
\quad single frame (no history) & 30.93 & $-$1.65 & 0.8796 & 0.170 & 0.0668 & 8.20 & 1.09 \\
\multicolumn{8}{@{}l}{QRISP, $\times$4} \\
\quad recurrent & 29.74 &  & 0.8531 & 0.196 & 0.0829 & 7.69 & 0.44 \\
%      (difference of the two numbers above)
\quad single frame (no history) & 27.71 & $-$2.02 & 0.7783 & 0.320 & 0.1074 & 6.68 & 1.74 \\
\multicolumn{8}{@{}l}{Arm Bistro, $\times$2, zero-shot} \\
\quad recurrent & 31.07 &  & 0.9506 & 0.040 & 0.0764 & 9.23 & --- \\
%      (difference of the two numbers above)
\quad single frame (no history) & 26.64 & $-$4.43 & 0.8726 & 0.126 & 0.1174 & 8.22 & --- \\
\bottomrule
\end{tabular*}
\par\smallskip
\begin{minipage}{\linewidth}\scriptsize
$\Delta$: PSNR-Y difference to bicubic for the Lanczos rows, to variance-clipping TAAU for the clamping rows, and to the recurrent model in (b). Differences are computed before rounding. (a) Bicubic and unqualified Lanczos-3 use the unjittered MipBias render; TAAU and rows marked `jittered input' use the jittered MipBiasJit render that the networks receive. (b) Recurrent: the analysis recipe ($\lambda_T=0.05$, 30-frame clips); single frame: the same network and recipe without history, accumulator or recurrent features. \reflectbox{F}LIP and JOD over the whole test set (Arm: its 92 scored frames); static std: computed after quantising the output to 8-bit values, AbandonedSchool/0001 frames 275--292, as in Table~\ref{tab:headline} of the paper; ---: not computed (the Arm capture has no static segment).
\end{minipage}
\end{table*}
%% ======== END tables/T8s_extra_rows ========

\paragraph*{Every test sequence and static segment at $\times$2.}
Table~\ref{supp:tab:x2detail} gives the per-sequence and per-segment numbers behind Section~\ref{sec:main} of the paper:
against DLSS~2 our model is higher in PSNR-Y and JOD and lower in \reflectbox{F}LIP on all five sequences, and on all
three static segments it is higher in JOD, lower in \reflectbox{F}LIP and lower in pixel standard deviation. On the two
SeaPort stops, pixel standard deviation is 1.05 versus 1.53 and 1.49 versus 1.76 for DLSS~2; the reference itself changes
over both stops.

The SeaPort reference changes during both camera stops.
A zero-variance output would not exactly match this reference sequence. The lower output variation is therefore assessed
with JOD and FLIP, which also favour our model on these segments, rather than by requiring the output variance
to vanish.

%% ======== BEGIN tables/T8s_x2_detail ========
% T8s_x2_detail.tex -- generated by paper/make_tables.py (2026-10-01 15:56 UTC); do not edit by hand
% except the caption, which is preserved when the table is regenerated.
% Every number is read from the file named in the '% src:' comment above its row.
\begin{table*}[t]
\centering
\caption{Ours and DLSS~2 at $\times$2 on every test sequence and every static segment. Ours is higher in PSNR-Y and JOD and lower in \reflectbox{F}LIP on all five sequences, and higher in JOD, lower in \reflectbox{F}LIP and lower in pixel standard deviation on all three static segments.}
\label{supp:tab:x2detail}
\footnotesize
\setlength{\tabcolsep}{3pt}
\begin{tabular*}{\textwidth}{@{\extracolsep{\fill}}l r r r r r r r r r@{}}
\toprule
 & \multicolumn{2}{c}{PSNR-Y$\,\uparrow$} & \multicolumn{2}{c}{\reflectbox{F}LIP$\,\downarrow$} & \multicolumn{2}{c}{JOD$\,\uparrow$} & \multicolumn{3}{c}{pixel std (8-bit)} \\
\cmidrule(lr){2-3}\cmidrule(lr){4-5}\cmidrule(lr){6-7}\cmidrule(l){8-10}
Sequence / segment & Ours & DLSS~2 & Ours & DLSS~2 & Ours & DLSS~2 & Ours & DLSS~2 & Ref. \\
\midrule
\multicolumn{10}{@{}l}{\textit{whole sequences (frames 8--299)}} \\
AbandonedSchool/0000 & 25.87 & 24.58 & 0.1137 & 0.1282 & 7.77 & 7.35 & --- & --- & --- \\
AbandonedSchool/0001 & 37.92 & 35.69 & 0.0384 & 0.0486 & 9.14 & 8.80 & --- & --- & --- \\
SeaPort/0000 & 30.04 & 27.39 & 0.0525 & 0.0694 & 8.40 & 7.92 & --- & --- & --- \\
SpaceShipDemo/0000 & 36.83 & 33.27 & 0.0400 & 0.0533 & 9.23 & 8.80 & --- & --- & --- \\
SpaceShipDemo/0001 & 35.43 & 32.40 & 0.0447 & 0.0705 & 8.71 & 8.51 & --- & --- & --- \\
\midrule
\multicolumn{10}{@{}l}{\textit{static segments (camera stops)}} \\
AbandonedSchool/0001, frames 274--292 & --- & --- & 0.0341 & 0.0366 & 8.74 & 8.58 & 0.13 & 0.45 & 0.01 \\
SeaPort/0000, stop 1, frames 20--38 & --- & --- & 0.0368 & 0.0460 & 8.55 & 8.12 & 1.05 & 1.53 & 0.98 \\
SeaPort/0000, stop 2, frames 205--223 & --- & --- & 0.0559 & 0.0650 & 7.47 & 7.20 & 1.49 & 1.76 & 1.29 \\
\bottomrule
\end{tabular*}
\par\smallskip
\begin{minipage}{\linewidth}\scriptsize
Ours: the $\times$2 model in Table~\ref{tab:headline} of the paper (100k iterations); DLSS~2: the shipped frames, scored with our harness. JOD: ColorVideoVDP; pixel std: per-pixel temporal standard deviation of the 8-bit output over the segment, averaged over the frame (Ref.: the reference, which changes on the SeaPort stops). The AbandonedSchool segment here spans frames 274--292 (the JOD$_{\mathrm s}$ segments); Table~\ref{tab:headline} uses 275--292 for its static std and static PSNR.
\end{minipage}
\end{table*}
%% ======== END tables/T8s_x2_detail ========

\paragraph*{R and J models per scene.}
Table~\ref{supp:tab:scenes} gives the per-scene PSNR of the $\times$4 grid models at $\lambda_T=0$ and $0.05$. The R
models trail the J models only on SpaceShipDemo (by 0.13 and 0.23\,dB); at $\lambda_T=0$ the whole deficit
lies in the sequence with the particle sprites of Figure~\ref{supp:fig:failure} (0001).

%% ======== BEGIN tables/T8s_scenes ========
% T8s_scenes.tex -- generated by paper/make_tables.py (2026-10-01 15:56 UTC); do not edit by hand
% except the caption, which is preserved when the table is regenerated.
% Every number is read from the file named in the '% src:' comment above its row.
\begin{table}[t]
\centering
\caption{Per-scene PSNR-Y of the R and J $\times$4 grid models (192$\times$8, 45k iterations), with the two SpaceShipDemo sequences.
}
\label{supp:tab:scenes}
\footnotesize
\setlength{\tabcolsep}{3pt}
\begin{tabular}{@{}l l r r r r r r@{}}
\toprule
 & & Abandoned- & & \multicolumn{3}{c}{SpaceShipDemo} & \\
\cmidrule(lr){5-7}
$\lambda_T$ & & School & SeaPort & all & 0000 & 0001 & Mean \\
\midrule
0 & R & 28.95 & 27.59 & 32.72 & 33.96 & 31.47 & 29.75 \\
 & J & 28.86 & 27.44 & 32.85 & 33.87 & 31.83 & 29.72 \\
%      (difference of the two numbers above)
%      (difference of the two numbers above)
%      (difference of the two numbers above)
%      (difference of the two numbers above)
%      (difference of the two numbers above)
%      (difference of the two numbers above)
 & R$-$J & +0.10 & +0.15 & $-$0.13 & +0.09 & $-$0.35 & +0.04 \\
\midrule
0.05 & R & 28.94 & 27.55 & 32.72 & 33.81 & 31.63 & 29.74 \\
 & J & 28.91 & 27.49 & 32.95 & 33.99 & 31.91 & 29.78 \\
%      (difference of the two numbers above)
%      (difference of the two numbers above)
%      (difference of the two numbers above)
%      (difference of the two numbers above)
%      (difference of the two numbers above)
%      (difference of the two numbers above)
 & R$-$J & +0.02 & +0.07 & $-$0.23 & $-$0.19 & $-$0.28 & $-$0.05 \\
\bottomrule
\end{tabular}
\par\smallskip
\begin{minipage}{\linewidth}\scriptsize
Scenes: mean over the scene's sequences; Mean: the test-set PSNR-Y (mean over the three scenes), as in Table~\ref{supp:tab:grid}. The particle sprites without motion vectors (Figure~\ref{supp:fig:failure}) are in SpaceShipDemo/0001.
\end{minipage}
\end{table}
%% ======== END tables/T8s_scenes ========

\paragraph*{\reflectbox{F}LIP on static and moving frames.}
Table~\ref{supp:tab:flip-split} splits the \reflectbox{F}LIP error at $\times$2 into the static-camera and the moving frames:
our model has the lowest error on both (0.046 and 0.059, against 0.055 and 0.075 for DLSS~2).

The static and moving columns are plain frame averages; the overall column gives each scene equal weight.
The subset sizes differ, so a scene-balanced score need not equal a frame-count-weighted average of the
subset scores.

%% ======== BEGIN tables/T8s_flip_split ========
% T8s_flip_split.tex -- generated by paper/make_tables.py (2026-10-01 15:56 UTC); do not edit by hand
% except the caption, which is preserved when the table is regenerated.
% Every number is read from the file named in the '% src:' comment above its row.
\begin{table}[t]
\centering
\caption{\reflectbox{F}LIP on static-camera and moving frames ($\times$2, QRISP test set).
}
\label{supp:tab:flip-split}
\footnotesize
\setlength{\tabcolsep}{4pt}
\begin{tabular}{@{}l r r r@{}}
\toprule
Method & static (97) & moving (1363) & all \\
\midrule
Ours (192$\times$8, 100k) & \textbf{0.0458} & \textbf{0.0587} & \textbf{0.0570} \\
DLSS 2 (shipped frames) & 0.0546 & 0.0754 & 0.0732 \\
TAAU (classical) & 0.0568 & 0.0760 & 0.0743 \\
Bicubic & 0.0593 & 0.0731 & 0.0719 \\
\bottomrule
\end{tabular}
\par\smallskip
\begin{minipage}{\linewidth}\scriptsize
\reflectbox{F}LIP (LDR, 67 pixels per degree), lower is better; bold: best. Static and moving: plain means over the static-camera frames and the moving frames (as in Table~\ref{supp:tab:split}); all: mean per scene, then over the three scenes, as in Table~\ref{tab:headline} of the paper.
\end{minipage}
\end{table}
%% ======== END tables/T8s_flip_split ========

\paragraph*{Clip length and the freeze.}
With 16-frame rather than 30-frame training clips in this 45k-iteration $\times$4 comparison, deployed static median
$\alpha$ falls from $2.3\times10^{-3}$ to $8.3\times10^{-4}$, indicating reduced uptake of new samples.

\paragraph*{int8 inference.}
With TensorRT, an int8 trunk costs 0.13\,dB PSNR for the 64-channel, 4-block network and 0.34 to 0.35\,dB for the
192-channel, 8-block network, averaged over three unbroken 300-frame passes, and up to 0.49\,dB on one scene
(Table~\ref{supp:tab:int8}); an fp16 trunk changes PSNR by less than 0.001\,dB.

\paragraph*{Seed pairs.}
Table~\ref{supp:tab:seeds} lists every pair of 15k-iteration runs whose configurations differ only in the seed. Across the
seven pairs, absolute PSNR differences range from 0.003 to 0.076\,dB, with a median of 0.026\,dB.

Each difference compares two runs with the same stated configuration. The accumulator initialisation in the
marked pair differs from the large-tier control of Table~\ref{supp:tab:design}, as its footnote records.
The observed range describes these seven pairs under the listed configurations.

%% ======== BEGIN tables/T8s_seeds ========
% T8s_seeds.tex -- generated by paper/make_tables.py (2026-10-01 15:56 UTC); do not edit by hand
% except the caption, which is preserved when the table is regenerated.
% Every number is read from the file named in the '% src:' comment above its row.
\begin{table}[t]
\centering
\caption{Seed pairs: every pair of 15k-iteration runs ($\times$4) whose configurations differ only in the seed.
}
\label{supp:tab:seeds}
\footnotesize
\setlength{\tabcolsep}{4pt}
\begin{tabular}{@{}l r r r r r@{}}
\toprule
Trunk & $\lambda_T$ & Clip & Seed 0 & Seed 1 & $|\Delta|$ \\
\midrule
% pair b11_plain_96x8 / b11_plain_96x8_s1
%      (absolute difference)
96$\times$8 & 0.05 & 30 & 29.164 & 29.168 & 0.003 \\
% pair b1_r / b1_r_s1
%      (absolute difference)
192$\times$8 & 0.25 & 16 & 28.818 & 28.827 & 0.009 \\
% pair b3_rtw05 / b4_rtw05_s1
%      (absolute difference)
192$\times$8 & 0.05 & 16 & 29.277 & 29.266 & 0.011 \\
% pair b4_rtw05_e / b6_rtw05_e_s1
%      (absolute difference)
192$\times$8$^{b}$ & 0.05 & 30 & 29.424 & 29.398 & 0.026 \\
% pair b11_rep3_96x8 / b11_rep3_96x8_s1
%      (absolute difference)
96$\times$8, ECB & 0.05 & 30 & 29.375 & 29.346 & 0.028 \\
% pair b1_rtw0 / b3_rtw0_s1
%      (absolute difference)
192$\times$8 & 0 & 16 & 29.312 & 29.376 & 0.064 \\
% pair b1_c0 / b1_c0_s1
%      (absolute difference)
192$\times$8$^{a}$ & 0.25 & 16 & 28.673 & 28.749 & 0.076 \\
% range 0.0034 to 0.0764 dB, median 0.0261 dB over 7 pairs
\bottomrule
\end{tabular}
\par\smallskip
\begin{minipage}{\linewidth}\scriptsize
PSNR-Y of the two seeds and their absolute difference, sorted: 7 pairs, 0.003 to 0.076\,dB, median 0.026\,dB. Registered model, 30- or 16-frame training clips as given. $^{a}$Only the unjittered motion field; the other registration pieces are off. ECB: re-parameterised trunk (Table~\ref{supp:tab:design}). $^{b}$These runs initialise the accumulator from the reference frame on 30\% of training clips; the large-tier control of Table~\ref{supp:tab:design} (29.48\,dB) does not.
\end{minipage}
\end{table}
%% ======== END tables/T8s_seeds ========

\begin{figure}[!htbp]
  \centering
  \includegraphics[width=\columnwidth]{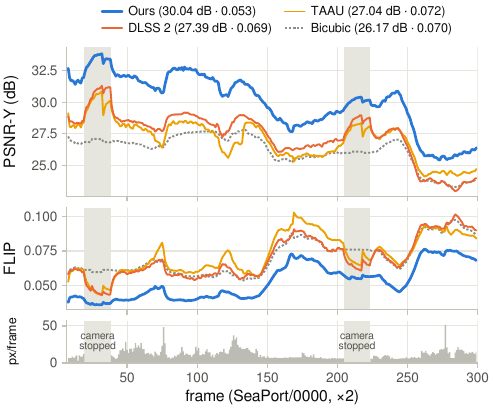}
  \caption{Per-frame PSNR-Y (top) and FLIP (middle) on SeaPort/0000 at $\times$2, scored frames 8--299:
ours, DLSS~2, TAAU and bicubic. Grey bands mark camera stops; the bottom strip shows median camera motion.
The legend gives sequence means.}
  \label{supp:fig:motion}
\end{figure}

\section{Qualitative gallery}\label{supp:section-16}\label{supp:sec:gallery}
Figures~\ref{supp:fig:gal-x2}--\ref{supp:fig:gal-x4} show two selected crops per scene at $\times$2, $\times$3 and $\times$4.
Candidate windows are $192\times96$ ($\times$2) or $240\times96$ pixels ($\times$3, $\times$4); a window is eligible
when its reference detail (box mean of $|\nabla^2 Y|$) is at least the larger of the frame's 75th percentile and 8,
its mean luma is at least 40, and our crop \reflectbox{F}LIP is below every baseline's. We select the window with the
largest PSNR-Y gain over the best baseline of that window. For AbandonedSchool and SpaceShipDemo, the second crop is the
best window from the other camera sequence; for SeaPort, which has a single test sequence, it is the best window at
least 30 frames away from the first. Figure~\ref{supp:fig:motion} follows one test
sequence frame by frame. The supplementary video and the interactive viewer show the same comparisons in motion.

The selection rule uses our crop PSNR-Y and FLIP, so these images illustrate locations where the method has
a large measured advantage. The complete-sequence measurements provide the average scores. The error maps use the same range within
the galleries, and each crop is compared with the
reference at the same location.

The per-frame comparison in Figure~\ref{supp:fig:motion} provides a different view. On all 292 scored SeaPort
frames, our model has higher PSNR-Y and lower FLIP than every plotted baseline. The minimum PSNR-Y margin
over the strongest baseline at each frame is 1.03\,dB, including frames immediately after the camera resumes
moving. On AbandonedSchool/0001 and SpaceShipDemo/0001, at least one baseline has equal or higher PSNR-Y on
2 and 3 frames. At least one baseline has lower FLIP on 17, 14 and 21 frames of AbandonedSchool/0001,
SpaceShipDemo/0000 and SpaceShipDemo/0001, respectively.

% References close the analytical text; the three full-page gallery plates follow.
\newpage

\FloatBarrier

\begin{galfigure}[p]
  \centering
  \includegraphics[width=\textwidth]{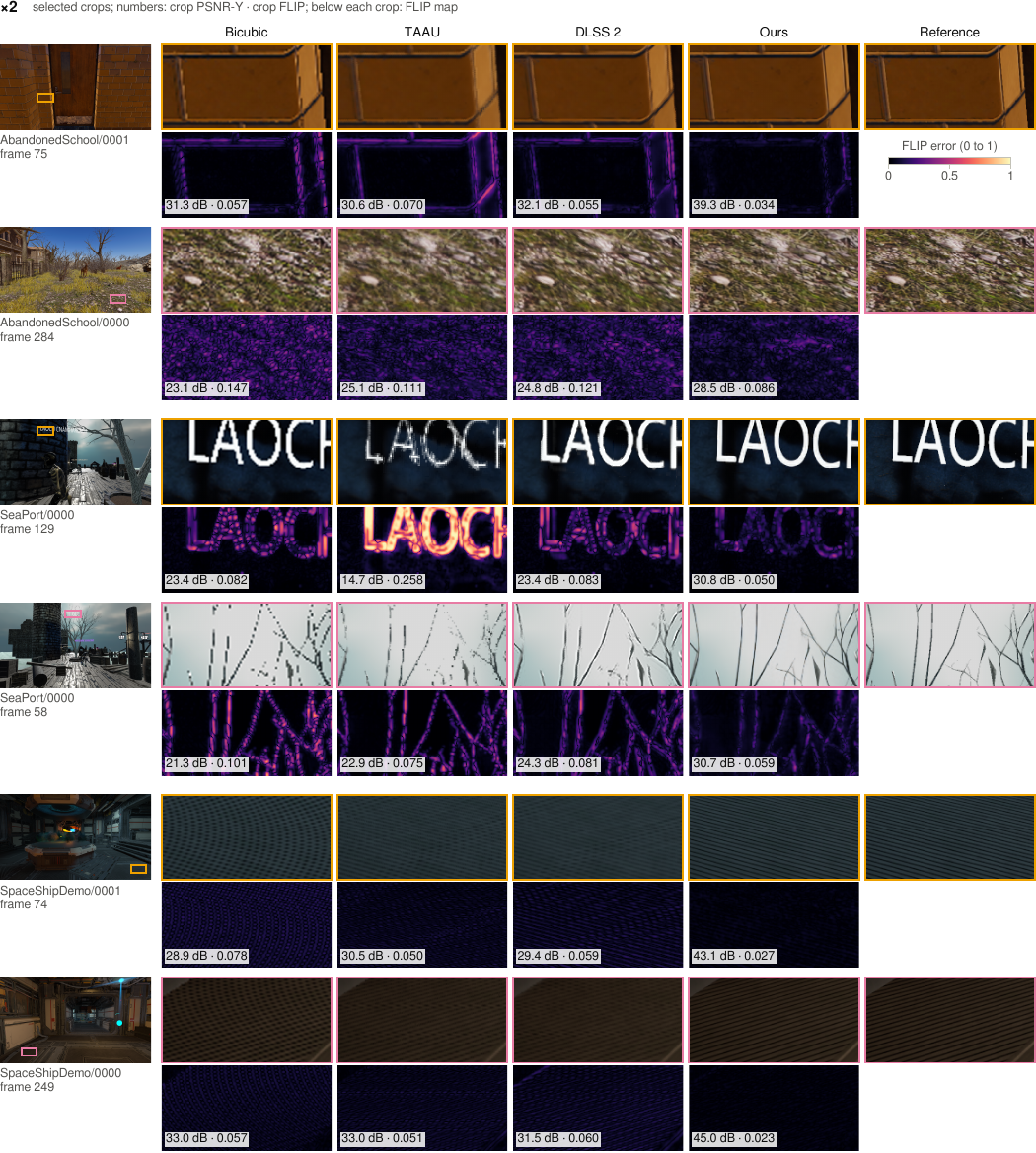}
  \caption{Gallery at $\times$2 (selected crops). Each row shows the full reference frame with the crop box, bicubic upsampling, our classical TAAU, DLSS~2 (shipped frames), our model and the reference; below each crop is its \reflectbox{F}LIP error map (0 to 1), labelled with crop PSNR-Y and \reflectbox{F}LIP. Our model recovers thin and periodic structure (branches, lettering, floor gratings) where the baselines alias, blur or break it up; in the SeaPort frame-129 crop, TAAU produces fragmented lettering and scores 14.7\,dB PSNR-Y.}
  \label{supp:fig:gal-x2}
\end{galfigure}

\begin{galfigure}[p]
  \centering
  \includegraphics[width=\textwidth]{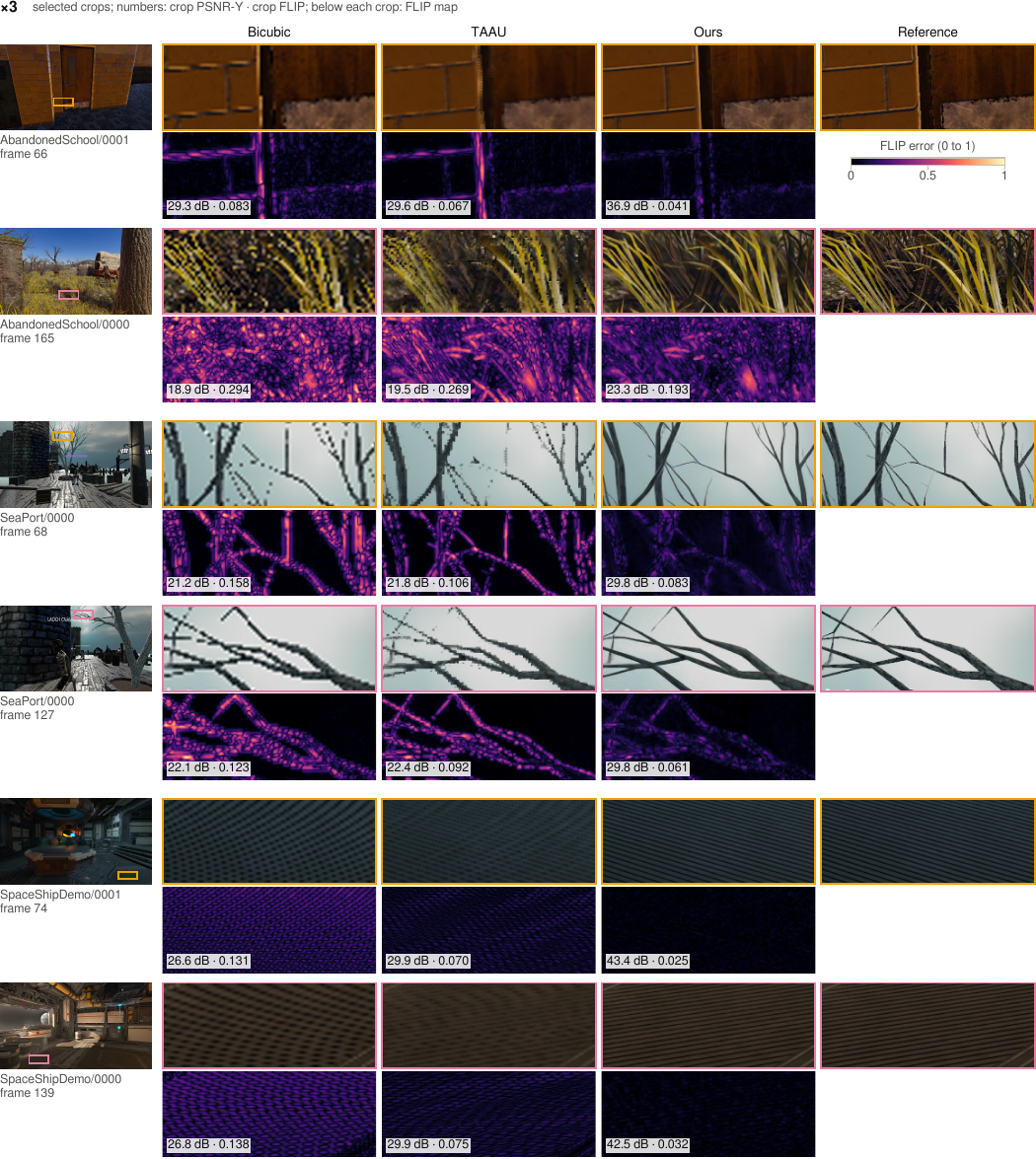}
  \caption{Gallery at $\times$3 (selected crops), as in Figure~\ref{supp:fig:gal-x2}; the QRISP dataset provides DLSS~2 frames only at $\times$2.}
  \label{supp:fig:gal-x3}
\end{galfigure}

\begin{galfigure}[p]
  \centering
  \includegraphics[width=\textwidth]{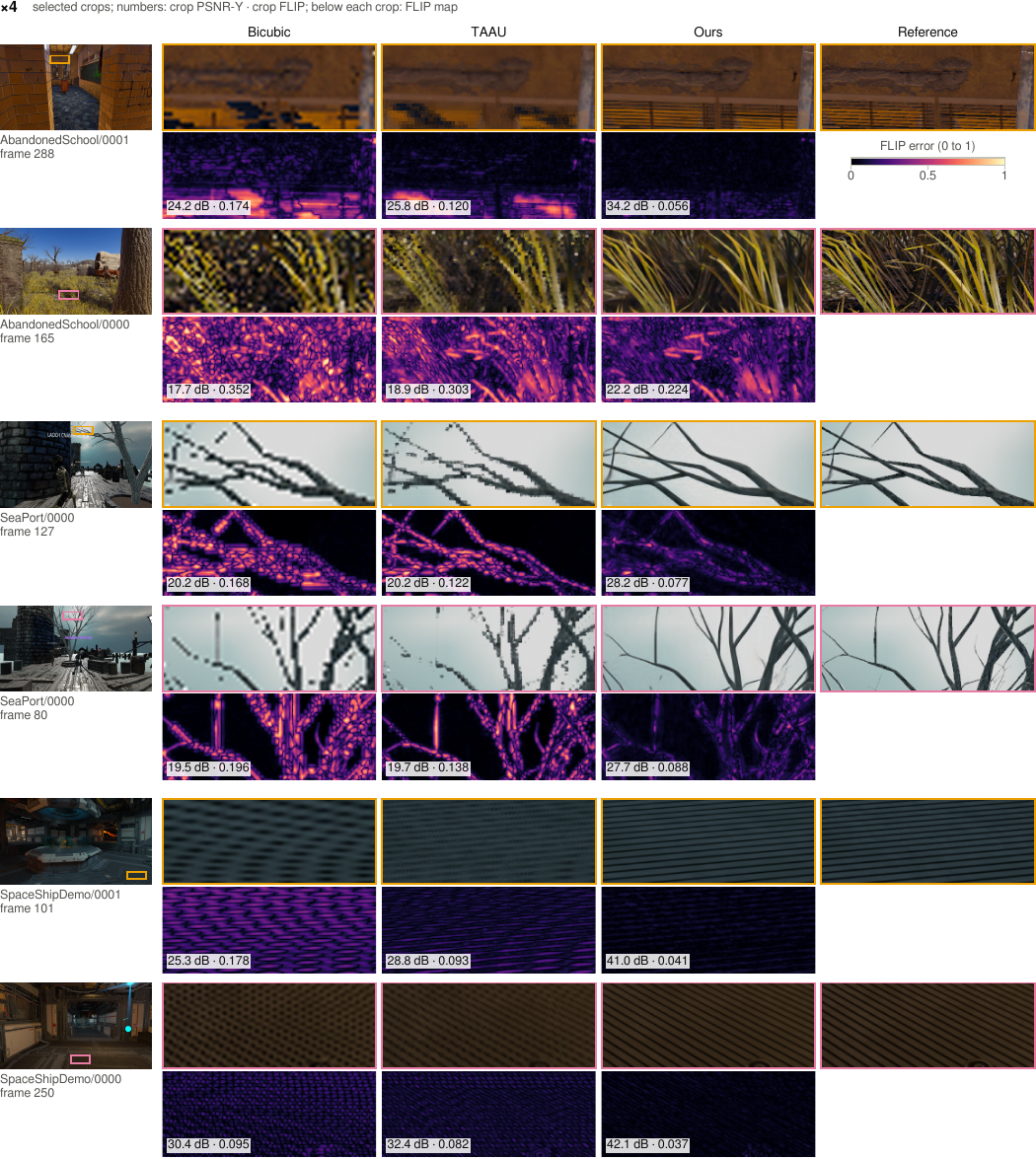}
  \caption{Gallery at $\times$4 (selected crops), as in Figure~\ref{supp:fig:gal-x2}. At $\times$4, the input has one low-resolution pixel per 16 output pixels. In these selected crops, our model reconstructs grass blades, branches and floor gratings with less blocking and moir\'e than bicubic upsampling and TAAU.}
  \label{supp:fig:gal-x4}
\end{galfigure}

\FloatBarrier

\end{document}